\documentclass[sigconf]{acmart}

\newcommand\ies{\textit{i.e.}}
\newcommand\egs{\textit{e.g.}}
\newcommand\etals{\textit{et al.}~}

\usepackage{color}
\usepackage{arydshln}
\usepackage{bm}
\usepackage{textcomp}
\usepackage{bbding}
\usepackage{multirow}
\usepackage{xcolor}
\definecolor{citecolor}{HTML}{0071bc}
\usepackage[export]{adjustbox} 
\AtBeginDocument{%
  \providecommand\BibTeX{{%
    \normalfont B\kern-0.5em{\scshape i\kern-0.25em b}\kern-0.8em\TeX}}}

\copyrightyear{2022}
\acmYear{2022}
\setcopyright{acmcopyright}
\acmConference[MM '22] {Proceedings of the 30th ACM International Conference on Multimedia}{October 10--14, 2022}{Lisboa, Portugal.}
\acmBooktitle{Proceedings of the 30th ACM International Conference on Multimedia (MM '22), Oct. 10--14, 2022, Lisboa, Portugal}
\acmPrice{15.00}
\acmISBN{978-1-4503-9203-7/22/10}

\acmDOI{10.1145/3503161.3547907}

\acmSubmissionID{654}

\begin{document}

\title{Joint Learning Content and Degradation Aware Embedding for Blind Super-Resolution}






\author{Yifeng Zhou$^{1\ast}$ \quad Chuming Lin$^{1\ast}$ \quad  Donghao Luo$^{1}$ \quad Yong Liu$^{1}$  \quad  Ying Tai$^{1\dagger}$ \and Chengjie Wang$^{1\dagger}$ \quad Mingang Chen$^{2}$}

\makeatletter
\def\authornotetext#1{
\if@ACM@anonymous\else
    \g@addto@macro\@authornotes{
    \stepcounter{footnote}\footnotetext{#1}}
\fi}
\makeatother
\authornotetext{Equal contribution. $^{\dagger}$Corresponding author.}

\affiliation{
 \institution{\textsuperscript{\rm 1}Youtu Lab, Tencent\country{China}}
 \institution{\textsuperscript{\rm 2}Shanghai Development Center of Computer Software Technology\country{China}}
 }
\email{{joefzhou, chuminglin, michaelluo, choasliu, yingtai, jasoncjwang}@tencent.com, cmg@sscenter.sh.cn}

\def\authors{Yifeng Zhou, Chuming Lin, Donghao Luo, Yong Liu, Ying Tai, Chengjie Wang, Mingang Chen}
\renewcommand{\shortauthors}{Yifeng Zhou, et al.}


\begin{abstract}
  To achieve promising results on blind image super-resolution (SR), 
  some attempts leveraged the low resolution (LR) images to predict the kernel and improve the SR performance. 
  However, these \textit{Supervised Kernel Prediction} (SKP) methods are impractical due to the unavailable real-world blur kernels. 
  Although some \textit{Unsupervised Degradation Prediction} (UDP) methods are proposed to bypass this problem, the \textit{inconsistency} between degradation embedding and SR feature is still challenging. 
  By exploring the correlations between degradation embedding and SR feature, we observe that jointly learning the content and degradation aware feature is optimal. 
  Based on this observation, a Content and Degradation aware SR Network dubbed CDSR is proposed. 
  Specifically, CDSR contains three newly-established modules: 
  (1) a Lightweight Patch-based Encoder (LPE) is applied to jointly extract content and degradation features; 
  (2) a Domain Query Attention based module (DQA) is employed to adaptively reduce the inconsistency; 
  (3) a Codebook-based Space Compress module (CSC) that can suppress the redundant information. 
  Extensive experiments on several benchmarks demonstrate that the proposed CDSR outperforms the existing UDP models and achieves competitive performance on PSNR and SSIM even compared with the state-of-the-art SKP methods.

\end{abstract}





\maketitle
\section{Introduction}
\begin{figure}[t]
	\centering
	\includegraphics[width=\linewidth]{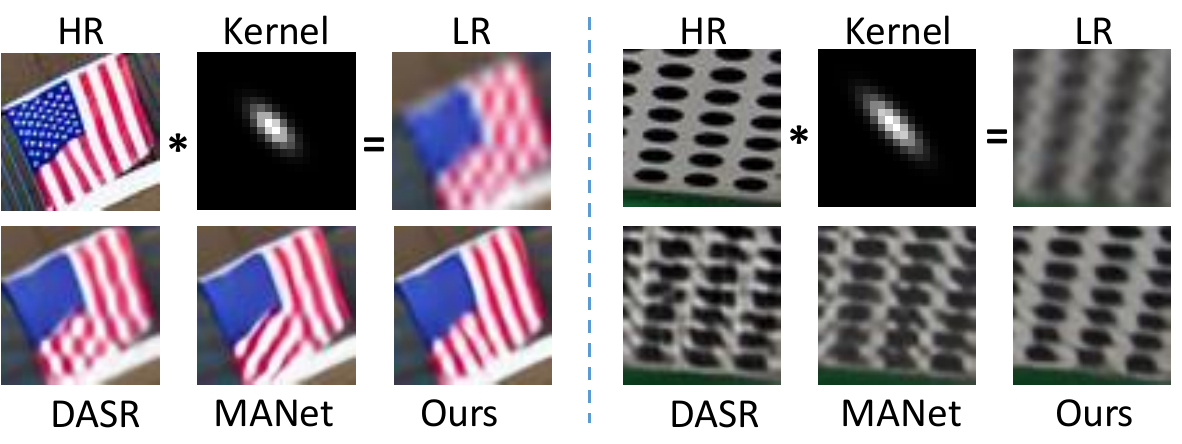}
	\vspace{-7mm}
	\caption{
		Some SR results for scale factor of $\times 4$. Existing methods produce artifacts due to they apply the moderate receptive field (MANet~\cite{manet2021}), implicit content-aware degradation embedding (DASR~\cite{Wang2021Unsupervised_dasr}). 
		We introduce the content information as the cue to enhance the cooperation between degradation embedding and the SR network.
	}
	\label{fig:teaser}
	\vspace{-5mm}
\end{figure}

Blind image super-resolution aims to restore high-resolution images from low resolution
inputs with unknown degradation factors. 
Unlike single image super-resolution (SISR) methods~\cite{dong2014learning,zhang2018image,dong2016accelerating,wang2018sft,singh2014srnoise} which are developed based on a pre-defined degradation process (\egs, bicubic downsampling).
Blind SR towards higher generalization and practicability.
Usually, blind SR is achieved by two steps: degradation estimation within the prior encoder, and the fusion of degradation prior and textural features within the SR network.
Based on the designs of degradation extraction, existing blind SR approaches can be divided into two groups:

(1) \textit{Supervised Kernel Prediction} (SKP). 
Most existing works~\cite{gu2019blind_ikc,huang2020unfolding,manet2021,zhang2018SRMDNF,zhang2021BSRGAN,tao2021spectrum,zhang2020deepUnfold,kim2021koalanet} employ the classical degradation model to represent the degradation.
Previous SKP methods leverage explicit~\cite{gu2019blind_ikc,kim2021koalanet,correction_filter,zhang2018SRMDNF,soh2020_kernel_metazeroshotSR,zhang2020deepUnfold} or implicit~\cite{bell2019blind_kernelGAN,luo2022DCLS} ways to predict blur kernels, and then employ the kernel stretching strategy to provide the degradation information for non-blind SR networks.
However, SKP methods are impractical due to the real-world kernels are unavailable. 
Once given complex degradation deviating from their training distribution, these methods inevitably lead to inferior results.
Furthermore, SKP can only handle the blur kernel-based degradation, and cannot be extended to other degradation (\egs, noise).

(2) \textit{Unsupervised Degradation Prediction} (UDP). 
Unlike the SKP methods that only consider kernel degradation, UDP~\cite{Wang2021Unsupervised_dasr,zhang2021blind_contrasreplearning} has developed a more suitable manner for real-world applications with unknown degradation.
Instead of requiring supervision from the ground-truth kernel label, UDP leverages the \textit{degradation embedding} which can naturally avoid the drawback of SKP.
Besides, the degradation embedding learned in UDP can be applied to represent not only the blur kernel but also other degradations (\egs, noise).
As the most representative approach, DASR~\cite{Wang2021Unsupervised_dasr} investigates the degradation representation in an unsupervised manner through contrastive learning.
Although DASR has outperformed some SKP methods~\cite{gu2019blind_ikc, bell2019blind_kernelGAN} within the easy degradation, there is still a gap between DASR and the latest SKP~\cite{manet2021,luo2022DCLS} in more complex scenarios. 

Towards achieving a more effective and competitive UDP, we firstly investigate the key point on:~\emph{What kind of degradation embedding is really needed for SR networks?}
Surprisingly, we observed that using a degradation-oriented embedding will fail to improve the network performance or even end up with poor results, more details in Sec.~\ref{sec:Accuracy the Better?}.
Based on the inverse-U relationship between classification accuracy and content information claimed in~\cite{kumar2022surprising_UACCPS}, a reasonable explanation is that in order to classify the degradation, the encoder only preserves the degradation-related feature and ignores the \textit{content information}.
As mentioned in~\cite{Wang2021Unsupervised_dasr,gu2019blind_ikc}, the straightforward interaction between degradation space and content space will introduce interference.
Without explicitly considering the content information, the degradation-oriented embedding is agnostic to the textural information which serves the SR network and then results in the artifacts, shown in Fig.~\ref{fig:teaser}.

In this work, we polish the UDP method through three aspects, and then produce more promising results:
(1) 
Since the content information can serve as the cue for the SR network, we derive that \textit{content-aware degradation} helps address the interference from the domain gap between degradation and content spaces. 
The previous methods which only employ a small receptive field or naive encoder may be stumbled by the inconsistency between the above two spaces because these embeddings do not make full use of the content information.
Inspired by the recent patch-based transformers~\cite{dosovitskiy2021an_VIT, patchallyounedd}, we propose a Lightweight Patch-based Encoder (LPE) to extract both the content and degradation aware features.
The LPE not only promotes the extraction of content information but also introduces the patch degradation consistency, thus being more competitive to model the global degradation.
(2) To adaptively fuse the predicted embedding into the SR network, 
we propose Domain Query Attention based module (DQA) to achieve the content-aware fusion. 
(3) Moreover, as claimed in~\cite{gu2019blind_ikc}, using PCA to project the kernel representation to low-dimension can make the network easier to learn the relationship between degradation and SR.
However, PCA can not be trained end-to-end with SR network, thus failing to learn the adaptive degradation embedding.
We extend this and introduce a Codebook-based Space Compress module (CSC) to limit the basis of feature space, thus reducing redundancy.

Specifically, we propose a \textit{Content and Degradation Aware SR Network}, termed CDSR, which enhances the UDP based blind SR by narrowing the domain gap between degradation embedding and SR content feature. 
Extensive experiments show that the proposed CDSR outperforms the existing UDP models and achieves competitive performance on PSNR and SSIM even compared with the SOTA SKP methods.
The main contributions of this paper are as follows:
\begin{itemize}
\item We first analyze the relation between content information and degradation embedding. Based on this, we proposed a lightweight patch-based encoder (LPE) to extract content-aware degradation embedding features.
\item We present a Domain Query Attention based module (DQA) to adaptively fuse the predicted content and degradation aware embedding into the SR network.
\item Inspired by PCA, we introduce a Codebook-based Space Compress module (CSC) to limit the basis of feature space.
\end{itemize}

\vspace{-2mm}
\section{Related Work}
\subsection{Non-blind SR}
Since SRCNN~\cite{dong2014learning} was first proposed to learn the mapping from LR to HR, the booming deep learning techniques are widely applied to SISR.
Various elaborate network architectures~\cite{dong2016accelerating,ledig2017photo_bicSR,dai2019secondorder_bicSR,haris2018bp_bicSR,zhang2018RCAN_bicSR,kim2016accurate_bicSR,kim2016deeplyrecu_bicSR,wang2018sft,tai2017image,tai2017memnet} and complex loss functions~\cite{johnson2016perceptualSR_bicSR,lugmayr2020srflow_bicSR,zhang2018unreasonable_bicSR} are then proposed.
Following the residual design, Kim~\etals~\cite{kim2016accurate_bicSR} propose a deep residual SR network.
Besides the residual learning strategy, Zhang~\etals~\cite{zhang2018RCAN_bicSR} apply the channel-wise attention. 
Dai~\etals~\cite{dai2019secondorder_bicSR} introduce the second-order version to achieve promising results.
Kim~\etals~\cite{kim2016deeplyrecu_bicSR} employ the DRCN to recursively refine the extraction feature.
However, these approaches are developed based on \textit{a pre-defined degradation process} (\egs, bicubic downsampling), which can hardly hold true when applied to real-world images.
Given more complex degradation deviating from their assumed type, these methods inevitably lead to inferior results.
To be more flexible, some SR networks are designed to address different degradations with given corresponding priority.
Specifically, Wang~\etals~\cite{wang2018sft} achieve impressive results by considering the textural information in the SR network and introducing the SFT layer.
Later, SRMD~\cite{zhang2018SRMDNF} takes the degradation as the additional input to super-resolve LR images under different kernels.
Xu~\etals~\cite{xu2020unified_udvd} incorporate dynamic convolution to achieve more flexible alternative.
Although the above approaches achieve promising results, the kernels (\ies, priorities) are not available when extended to real-world images. 

\vspace{-2mm}
\subsection{Blind SR}
\noindent\textbf{Supervised Kernel Prediction.} 
In order to polish the non-blind SR networks, some attempts are leveraged to estimate the kernels, then employing some adaptive fusion strategies (\egs, AdaIN~\cite{huang2017_adain,wang2018sft}, kernel stretching~\cite{zhang2018SRMDNF}, dynamic blocks ~\cite{jia2016dynamicConv}).
The non-blind SR networks are sensitive to the kernel information consequently the optimal kernel estimation is important.
Gu \etals ~\cite{gu2019blind_ikc} proposed iterative kernel correction (IKC) to refine the predicted kernel progressively by observing the previous SR results. 
Furthermore, DAN ~\cite{luo2021endtoendDAN} has unfolded the previous iterating process into an end-to-end manner by corporately using the Restorer and Estimator and yields better results. 
MANet ~\cite{manet2021} leverages moderate receptive field and exploits channel interdependence to conduct the kernel estimation. 
KOALANet ~\cite{kim2021koalanet} employs the pixel-wised local kernels and dynamic filter to integrate kernel information for SR. 
Later, Luo \etals ~\cite{luo2022DCLS} propose DCLS module to generate clean features based on the reformulation and estimated kernel. 
These methods can achieve remarkable performance when given the ground-truth blur kernel.
However, in the real-world images, the blur kernels predicted by kernel-estimating methods deviate from those of the ground-truth.
SKP methods are sensitive to kernel estimation and cannot be extended to other degradations.

\noindent\textbf{Unsupervised Degradation Prediction.}
UDP has developed a more suitable manner for real-world applications with unknown degradation.
Instead of requiring the supervision from the ground-truth kernel label, UDP leverages the \textit{degradation embedding} which can naturally avoid the drawback of SKP.
Wang~\etals firstly propose DASR~\cite{Wang2021Unsupervised_dasr} to achieve unsupervised degradation prediction, using contrastive loss for unsupervised degradation representation learning.
In order to improve the contrastive learning procedure, Zhang~\etals~\cite{zhang2021blind_contrasreplearning} introduce CRL-SR to extract refined contrastive features via a bidirectional contrastive loss.
However, previous UDP approaches pay little attention to what kind of degradation embedding is really needed for blind SR.
In fact, they ignore that discriminative embedding which lacks content features will cause the problem of domain gap~\cite{gu2019blind_ikc}.
Instead, our proposed LPE jointly learns the content and degradation aware embedding, thus achieving promising results.
Besides, DQA can leverage the content feature to query the adaptive embedding feature for the SR network.

\subsection{Contrastive Learning}
Contrastive learning is prevalently studied for unsupervised representation learning.
Several recent studies~\cite{chen2020simple_CL,tian2020contrastive_CL,noroozi2017representation_CL,he2020moco} present promising results by minimizing the distance of similar tokens meanwhile maximizing that of dissimilar ones.
Specifically, some of the UDP methods leverage the contrastive learning approaches to distinguish the latent degradation from other ones.
However, to classify the degradation, the encoder only preserves the degradation-related feature and ignores the \textit{content information} therefore introducing the problem of \textit{domain gap}.
Although, previous methods have applied the effective contrastive learning approaches~\cite{he2020moco} to enrich the learning samples, there is also the redundancy between different degradations.
The proposed CSC can alleviate this by limiting the basis of feature space.

\begin{table}[!t]
	\caption{Classification accuracy of degradation embedding and PSNR of $\times 4$ SR. We choose the positive sample in three different ways. `{C\&D}': cropped from the same image with the same degradation (used in DASR). `{D}': cropped from different images with the same degradation.
	`{C}': cropped from the same image but with different degradation. 
	More discriminative embedding fails to achieve higher PSNR.}
	\label{tab:Accuracy the Better?}
	\vspace{-3mm}
	\begin{tabular}{c|c|cccc}
		\hline
		Positive  & Acc. & Set5         & Set14        & B100       & U100     \\ \hline
		\hline
		C\&D & 75.40\%          & 31.45 dB & 28.12 dB & 27.24 dB & 25.28 dB \\ \hline
		D    & 80.80\%          & 31.36 dB & 28.03 dB & 27.20 dB & 25.12 dB \\ \hline
		C    & 12.00\%          & 31.32 dB & 28.06 dB & 27.24 dB & 25.26 dB \\ \hline
	\end{tabular}
	\vspace{-5mm}
\end{table}

\section{Analysis on Degradation Embedding}
\label{sec:Accuracy the Better?}

\noindent\textbf{The Higher Degradation Accuracy the Better?} 
We firstly investigate the key point on: What kind of degradation embedding is really needed for SR networks?
As shown in Table~\ref{tab:Accuracy the Better?}, we conduct three experiments based on the SOTA DASR~\cite{Wang2021Unsupervised_dasr} to further investigate the learned degradation representation.
In Table~\ref{tab:Accuracy the Better?}, we show the effect of content information by only changing the positive selection strategy during contrastive learning.

Given $100$ images from B$100$~\cite{B100} and the pre-trained encoder $E(\cdot)$.
Each HR is degraded by $10$ different degradations to obtain the LR, using Anisotropic Gaussian kernels with $\sigma_1=\sigma_2 \in [1, 10]$ and $\theta$ as $0$. 
The scale factor is $\times 4$.
The average of $50$ sampled LR embedding from the same degradation is regarded as the cluster center $D$.
The rest of $50$ LR images $I$ are utilized to test the accuracy of classification, which can be mathematically calculated by:
\begin{equation}
Acc. = \frac{\sum_{i,j} Sign\{ \arg\max_k( Sim (E(I_i^j), D_k )), j \}}{N},
\end{equation}
where $i\in [1,50]$, $j,k \in [1, 10]$, $N=500$, $I_i^j$ denotes $i^{th}$ image which is degraded by $\sigma_1=\sigma_2=j$, $D_k$ denotes the $k^{th}$ cluster center, and $Sign(x,y)=1$ when $x = y $; $Sign(x,y)=0$ when $x\neq y$, the $Sim(x,y)$ calculates the cosine similarity of $x$ and $y$.
The accuracy denotes the encoder's capability of degradation classification.
In DASR ~\cite{Wang2021Unsupervised_dasr}, patches from the same image with the same degradation are considered as positive samples, while different degradations as negative counterparts, as shown in the first row of Table~\ref{tab:Accuracy the Better?}.
Based on contrastive learning, the encoder in DASR can capture the degradation information and implicitly learn the content information.

To enhance the capability of degradation classify (second row of Table~\ref{tab:Accuracy the Better?}), the network is constrained to focus on the degradation by discarding the implicit content information within patches from a single image.
Although it can achieve higher classification accuracy, the PSNR drops.
The reason may be the encoder only preserves the degradation-related feature and ignores the content information, as demonstrated in~\cite{kumar2022surprising_UACCPS} that there exists an inverse-U relationship between classify accuracy and content information.
Lacking content information introduces the domain gap between embedding and textural spaces, resulting in poor SR performance.

\begin{figure}[tb]
	\begin{tabular}[t]{c@{ }c@{ }}
		\includegraphics[width=.5\linewidth,height=.48\linewidth]{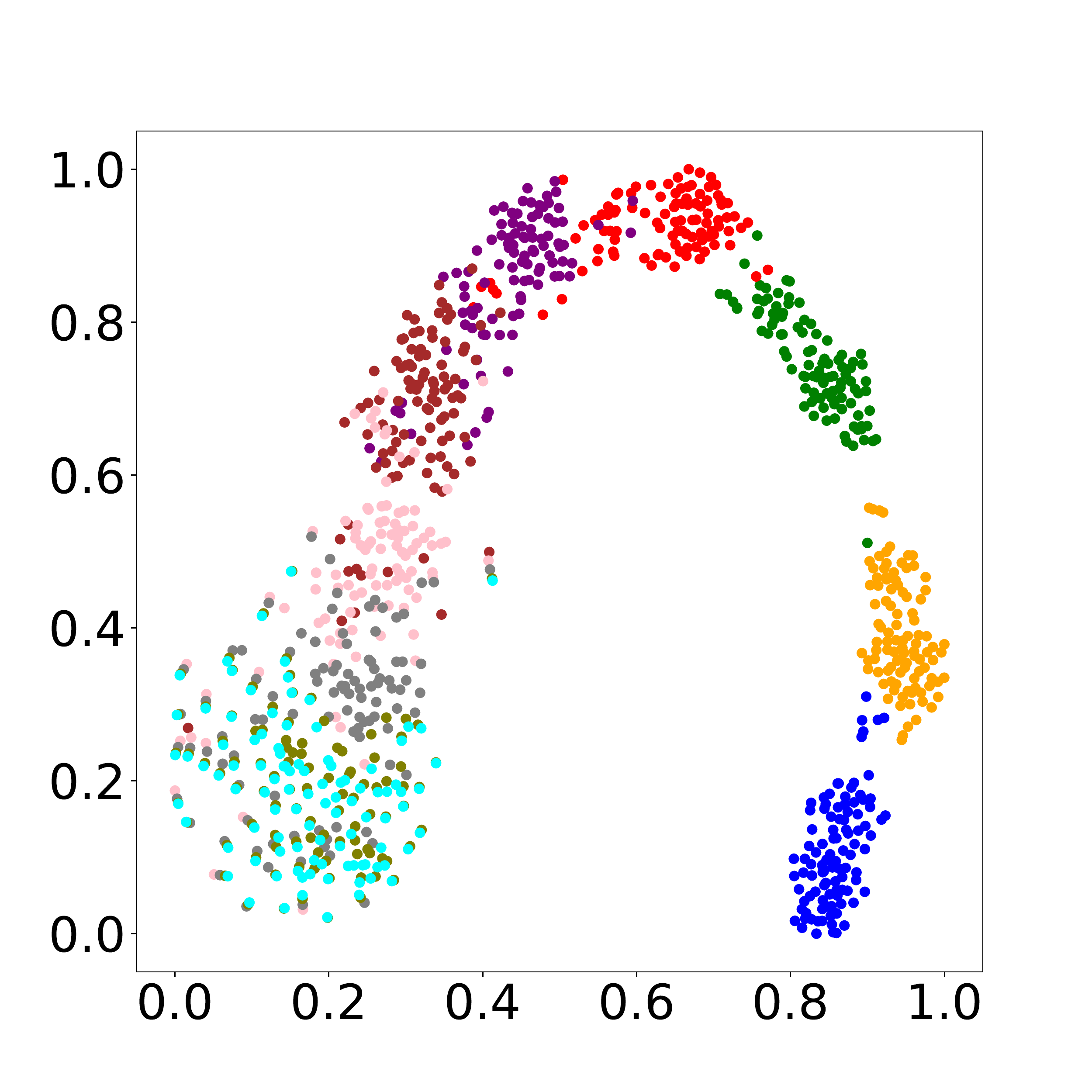}  &  \hspace{-2mm}
		\includegraphics[width=.5\linewidth,height=.48\linewidth]{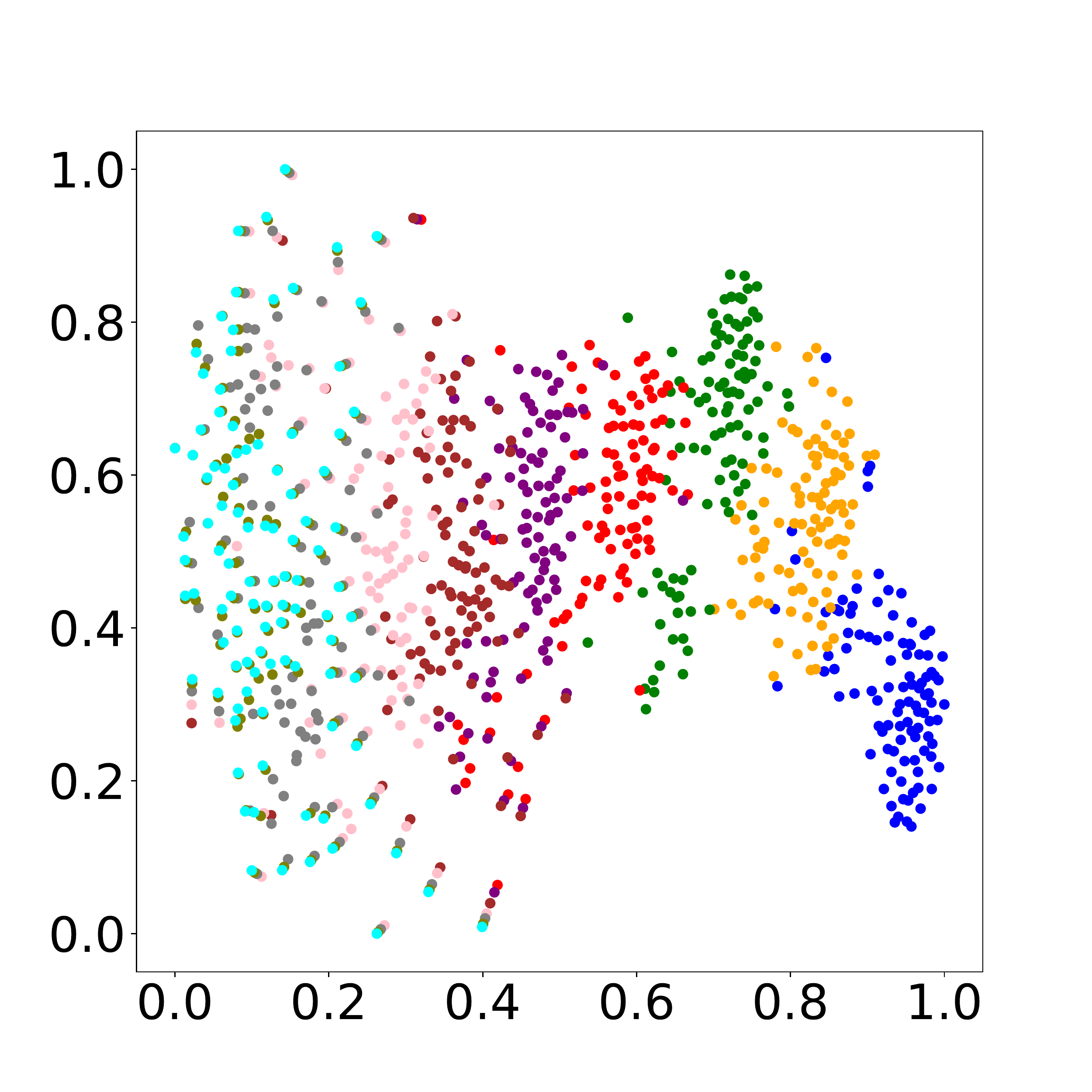}
		\vspace{-2mm}
		\\
		\includegraphics[width=.5\linewidth]{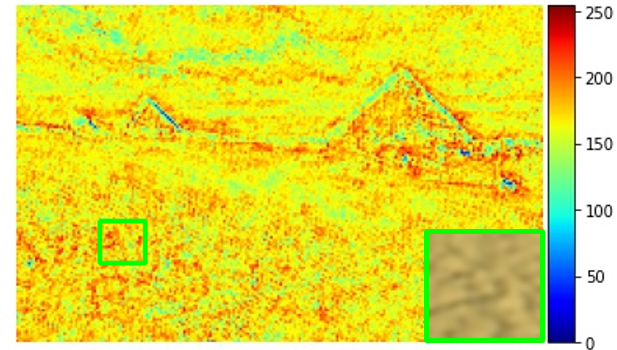}  &  \hspace{-2mm}
		\includegraphics[width=.5\linewidth]{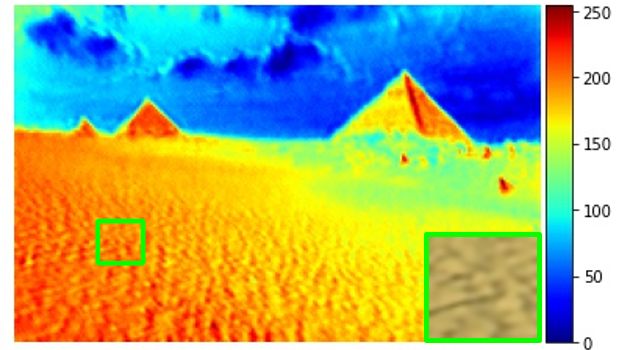}
		\\
		DASR  &  Ours 
	\end{tabular}
	\vspace{-4mm}
	\caption{Top: Representations for embedding with different blur kernels (denoted in colors). 
	Bottom: SR Feature visualization.
	DASR pays more attention to local degradation. 
	However, CDSR jointly learns the content and degradation aware embedding to achieve better results. $0\sim255$ denotes the active intensity of the feature map.}
	\label{fig:Accuracy the Better?}
	\vspace{-4mm}
\end{figure}

\begin{figure*}[!htb]
	\centering
	\includegraphics[width=.88\linewidth]{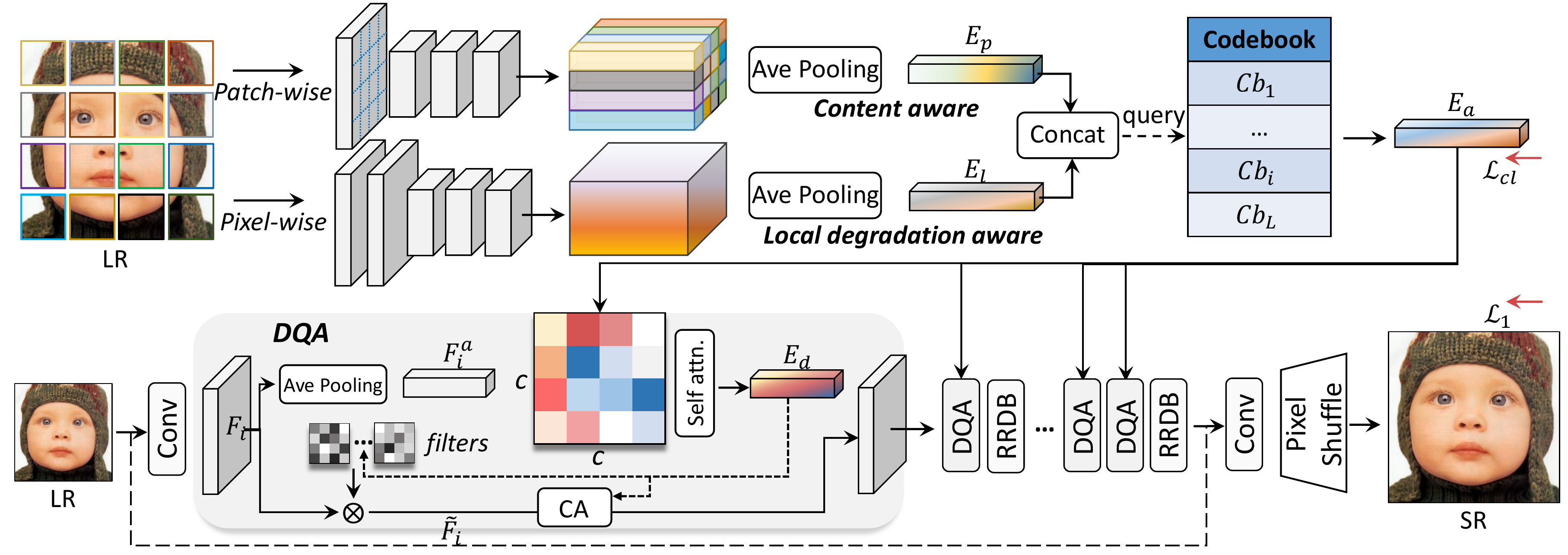}
	\vspace{-4mm}
	\caption{
		Top: Lightweight Patch-based Encoder (LPE) and Codebook-based Space Compression module (CSC).  LPE includes patch-wise feature extraction and pixel-wise feature extraction. 
		Bottom: SR network composed of proposed DQAs.
		The output content and degradation aware embedding $\bm{E_a}$ is utilized in each DQA to adaptively produce dynamic filters. 
	    Self-attention and channel attention modules (CA) are also applied in DQA to help improve performance.
	    }
	\label{fig:framework}
	\vspace{-3mm}
\end{figure*}

\noindent\textbf{Effect of Content-aware Embedding.} 
We then conduct the experiment where patches from the same image but different degradations are regarded as positive counterparts, the third row of Table~\ref{tab:Accuracy the Better?}.
Surprisingly, although the classification accuracy only occupies 12\%, it can achieve promising PSNR results, especially in the tough test sets \ies, B$100$ and U$100$.
Conclusively, the content information can narrow the domain gap between degradation and textural spaces.
When the encoder is guided to extract the \textit{degradation embedding} meanwhile preserving the \textit{content information}, it can produce the most suitable embedding for the SR network.
We use the t-SNE~\cite{tsne} to visualize the embedding space (see Fig.~\ref{fig:Accuracy the Better?}).
DASR produces better degradation classification results with larger inter distance but finally gets blurry results compared to our CDSR, which pays more attention on content information (\egs, pyramid and sand).
It indicates that more explicit content features might slightly disturb degradation classification. 
However, it can serve as a cue to recover complex textures and boost SR performance.

\section{Proposed Method}

\subsection{Lightweight Patch-based Encoder}
Unlike previous approaches~\cite{Wang2021Unsupervised_dasr, manet2021} that employ the naive or small receptive field encoder, we introduce the Lightweight Patch-based Encoder (LPE) to produce the content and degradation aware embedding.
LPE has two sub-modules which are designed to extract corresponded \textit{content} and \textit{local degradation-aware} features:

\noindent\textbf{Patch-wise Feature Extraction.}   
Inspired by the patch-based vision transformers~\cite{dosovitskiy2021an_VIT,patchallyounedd}, a simple lightweight patch-based encoder is designed to fast expand the receptive field of convolution and extract the content information.
Specifically, the patch embedding layer is utilized to replace the first standard convolution layer.
Then, the output patch feature $F_p \in \mathbb{R}^{\frac{H}{P} \times \frac{W}{P} \times C}$ is processed by several standard convolution layers.
Finally, we average the $F_p$ within the spatial dimension to get $E_p$.

\noindent\textbf{Pixel-wise Feature Extraction.}   
As claimed in~\cite{manet2021}, a moderate receptive field can help the network to estimate the blur kernel. 
Hence, the receptive field of this sub-module is fixed to a moderate range and then extracts the local degradation-related information.
Finally, the output is averaged to obtain $E_l$.

\subsection{Codebook-based Space Compression}
The previous SKP usually employs PCA for the reduction of feature space.
More experiments results~\cite{gu2019blind_ikc,zhang2018SRMDNF} demonstrated that instead of directly using the kernel, the stretched kernel representation can make it easier for the network to learn the relationship between degradation and SR.
To extend this advantage in UDP, we propose the Codebook-based Space Compression module (CSC).
The inclusion relationship in degradation introduces redundancy and expands the domain gap between degradation and SR feature space.
After feature extraction, the CSC is utilized to constrict the basis of embedding space.
Mathematically, this procedure can be described as:
\begin{equation}
\begin{aligned}
\bm{Q_e} &= MLP(Cat(\bm{E_p}, \bm{E_l})), \\
\bm{K_e} &= MLP(\bm{Cb}), \\
\bm{E_a} &= softmax(\bm{Q_e} \cdot \bm{K_e}^T) \cdot \bm{Cb},
\end{aligned}
\end{equation}
where $\bm{Cb} \in \mathbb{R}^{L \times C}$ denotes the codebook, $L$ denotes the length of codebook, and $C$ is the channel number of embedding. 

\begin{table*}[t]
	\caption{Quantitative comparison on different test sets with anisotropic Gaussian kernels. The best two results are highlighted in {\color{red}red} and {\color{blue}blue} colors, respectively. The supervised tag denotes whether the corresponding method belongs to SKP or UDP.}
	\label{tab:main_comp}
	\vspace{-3mm}
	\small
	\begin{tabular}{c|ccc|cccc}
		\hline
		\multicolumn{1}{c|}{Method} & \multicolumn{1}{c|}{Scale Factor} & \multicolumn{1}{c|}{Noise Level} & \multicolumn{1}{c|}{Supervised} & \multicolumn{1}{c|}{Set5} & \multicolumn{1}{c|}{Set14} & \multicolumn{1}{c|}{BSD100} & \multicolumn{1}{c}{Urban100} \\ \hline
		
		{\color{gray}SRSVD}~\cite{SRVSD}                       & $\times2$                                & 0                                & \Checkmark                                               & {\color{gray}34.51/0.8787}             & {\color{gray}31.10.0.8581}               & {\color{gray}29.71/0.7993}                & {\color{gray}28.08/0.7965}                  \\
		{\color{gray}IKC}~\cite{gu2019blind_ikc}                         & $\times2$                                & 0                                & \Checkmark                                               &{\color{gray}35.30/0.9381}              & {\color{gray}31.48/0.8797}               & {\color{gray}30.50/0.8545}                & {\color{gray}28.62/0.8689}                 \\
		{\color{gray}MANet}~\cite{manet2021}                        & $\times2$                                & 0                                & \Checkmark                                               & {\color{blue} \textbf{35.98/0.9420}} &  {\color{blue} \textbf{31.95}}{\color{red} \textbf{/0.8845}} & {\color{blue} \textbf{30.97}}{\color{red} \textbf{/0.8651}} & {\color{red} \textbf{29.61/0.8880}}                 \\
	    
		\cdashline{1-8}[3pt/2.5pt]
		
		HAN~\cite{HAN}                          & $\times2$                                & 0                                & \XSolid                                                & 26.83/0.7919              & 23.21/0.6888               & 25.11/0.6613                & 22.42/0.6571                  \\
		DIP~\cite{DIP}                          & $\times2$                                & 0                                & \XSolid                                                & 28.19/0.7939              & 25.66/0.6999               & 25.03/0.6762                & 22.97/0.6737                  \\
		KernelGAN + ZSSR~\cite{bell2019blind_kernelGAN}             & $\times2$                                & 0                                & \XSolid                                                & -                         & 23.92/0.6898               & 25.28/0.6395                & 21.97/0.6582                  \\
		HAN + Correction~\cite{HANcorrection}            & $\times2$                                & 0                                & \XSolid                                                & 28.61/0.8013              & 26.22/0.7292               & 26.88/0.7116                & 25.31/0.7109                  \\
		DASR~\cite{Wang2021Unsupervised_dasr}                         & $\times2$                                & 0                                & \XSolid                                                & 35.30/0.9360              & 31.30/0.8683               & 30.46/0.8507                & 28.66/0.8654                  \\
		Ours                         & $\times2$                                & 0                                & \XSolid                                                & {\color{red} \textbf{36.17/0.9428}} &  {\color{red} \textbf{32.14}}{\color{blue} \textbf{/0.8841}} &  {\color{red} \textbf{31.02}}{\color{blue} \textbf{/0.8643}}  & {\color{blue} \textbf{29.57/0.8851}}       \\
		
		\hline
		\hline
		
		{\color{gray}IKC}~\cite{gu2019blind_ikc}                          & $\times3$                                & 0                                & \Checkmark                                               & {\color{gray}32.94/0.9104}              & {\color{gray}29.14/0.8162}                & {\color{gray}28.36/0.7814}                & {\color{gray}26.34/0.8049}                  \\
		{\color{gray}MANet}~\cite{manet2021}                        & $\times3$                                & 0                                & \Checkmark                                               & {\color{blue} \textbf{33.69/0.9184}} & {\color{blue} \textbf{29.81/0.8270}} & {\color{blue} \textbf{28.81}}{\color{red} \textbf{/0.7932}} & {\color{blue} \textbf{27.39}}{\color{red} \textbf{/0.8331}}                \\
		
		\cdashline{1-8}[3pt/2.5pt]
		
		HAN~\cite{HAN}                          & $\times3$                                & 0                                & \XSolid                                                & 23.71/0.6171              & 22.31/0.5878               & 23.21/0.5653                & 20.34/0.5311                  \\
		DIP~\cite{DIP}                          & $\times3$                                & 0                                & \XSolid                                                & 27.51/0.7740              & 25.03/0.6674               & 24.60/0.6499                & 22.23/0.6450                  \\
		DASR~\cite{Wang2021Unsupervised_dasr}                         & $\times3$                                & 0                                & \XSolid                                                & 33.43/0.9151              & 29.57/0.8187               & 28.58/0.7846                & 26.83/0.8174                  \\
		Ours                         & $\times3$                                & 0                                & \XSolid                                                & {\color{red} \textbf{33.81/0.9192}} & {\color{red} \textbf{29.95/0.8275}} &  {\color{red} \textbf{28.81}}{\color{blue} \textbf{/0.7922}} &  {\color{red} \textbf{27.44}}{\color{blue} \textbf{/0.8329}}            \\
		
		\hline
		\hline
		
		{\color{gray}IKC}~\cite{gu2019blind_ikc}                          & $\times4$                                & 0                                & \Checkmark                                               & {\color{gray}31.08/0.8781}              & {\color{gray}27.83/0.7663}               & {\color{gray}27.12/0.7233}                & {\color{gray}25.16/0.7609}                  \\
		{\color{gray}MANet}~\cite{manet2021}                        & $\times4$                                & 0                                & \Checkmark                                               & {\color{blue} \textbf{31.54/0.8876}} & {\color{blue} \textbf{28.28/0.7727}} & {\color{blue} \textbf{27.36/0.7307}} & {\color{blue} \textbf{25.66/0.7760}}                  \\
		
		\cdashline{1-8}[3pt/2.5pt]
		
		HAN~\cite{HAN}                          & $\times4$                                & 0                                & \XSolid                                                & 21.71/0.5941              & 20.42/0.4937               & 21.48/0.4901                & 19.01/0.4676                  \\
		DIP~\cite{DIP}                          & $\times4$                                & 0                                & \XSolid                                                & 26.71/0.7417              & 24.52/0.6360               & 24.34/0.6160                & 21.85/0.6155                  \\
		KernelGAN + ZSSR~\cite{bell2019blind_kernelGAN}             & $\times4$                                & 0                                & \XSolid                                                & -                         & -                          & 18.24/0.3689                & 16.80/0.3960                  \\
		HAN + Correction~\cite{HANcorrection}              & $\times4$                                & 0                                & \XSolid                                                & 24.31/0.6357              & 24.44/0.6341               & 24.01/0.6005                & 22.32/0.6368                  \\
		DASR~\cite{Wang2021Unsupervised_dasr}                         & $\times4$                                & 0                                & \XSolid                                                & 31.45/0.8859              & 28.12/0.7703               & 27.24/0.7284                & 25.28/0.7636                  \\
		Ours                         & $\times4$                                & 0                                & \XSolid                                                & {\color{red} \textbf{31.63/0.8885}} & {\color{red} \textbf{28.31/0.7746}} & {\color{red} \textbf{27.38/0.7311}} & {\color{red} \textbf{25.75/0.7783}}                  \\
		
		\hline
		\hline
		
		{\color{gray}IKC}~\cite{gu2019blind_ikc}                          & $\times4$                                & 15                               & \Checkmark                                               & {\color{gray}27.23/0.7877}              & {\color{gray}25.55/0.6717}               & {\color{gray}25.15/0.6236}                & {\color{gray}23.31/0.6697}                  \\
		{\color{gray}MANet}~\cite{manet2021}                        & $\times4$                                & 15                               & \Checkmark                                               & {\color{blue} \textbf{27.58/0.7915}} & {\color{blue} \textbf{25.75/0.6744}} & {\color{blue} \textbf{25.30/0.6262}} & {\color{blue} \textbf{23.57/0.6760}}                  \\
		
		\cdashline{1-8}[3pt/2.5pt]
		
		HAN~\cite{HAN}                           & $\times4$                                & 15                               & \XSolid                                                & 20.88/0.4245              & 18.91/0.2901               & 21.01/0.4881                & 19.31/0.3552                  \\
		DIP~\cite{DIP}                          & $\times4$                                & 15                               & \XSolid                                                & 18.60/0.2695              & 18.14/0.2392               & 17.90/0.2073                & 18.82/0.3476                  \\
		KernelGAN + ZSSR~\cite{bell2019blind_kernelGAN}             & $\times4$                                & 15                               & \XSolid                                                & -                         & -                          & 19.56/0.4582                & 13.65/0.1136                  \\
		HAN + Correction~\cite{HANcorrection}             & $\times4$                                & 15                               & \XSolid                                                & 19.21/0.2281              & 18.21/0.2478               & 19.25/0.4231                & 19.01/0.3500                  \\
		DASR~\cite{Wang2021Unsupervised_dasr}                         & $\times4$                                & 15                               & \XSolid                                                & 27.48/0.7907              & 25.56/0.6723               & 25.25/0.6261                & 23.30/0.6663                  \\
		Ours                         & $\times4$                                & 15                               & \XSolid                                                & {\color{red} \textbf{27.70/0.7947}} & {\color{red} \textbf{25.81/0.6757}} & {\color{red} \textbf{25.33/0.6277}} & {\color{red} \textbf{23.60/0.6761}}              \\
		\hline
	\end{tabular}
\end{table*}

\subsection{Domain Query Attention based Module}
Existing methods usually fuse embeddings into non-blind SR networks without considering the domain gap.
The degradation features are different from the textural features which propagate in the SR network.
Therefore, we propose the Domain Query Attention based module to mitigate the domain gap.
Since $\bm{E_a}$ contains the content information, an intuitive solution is to leverage content cue to query the target value in $\bm{E_a}$.
Generally, each DQA employs the self-attention module to attend the embedding based on the current input feature $\bm{F_i}$.
Specifically, we first average the $\bm{F_i}$ within the spatial dimension to get $\bm{F_i^a} \in \mathbb{R}^{C \times 1}$.
Then, the channel-wise mutual self-attention is conducted in the DQA:
\begin{equation}
\label{eq:sa}
Attention(\bm{Q_d},\bm{K},\bm{V}) = softmax(\frac{\bm{Q_d}\bm{K}^T}{\sqrt{d_k}}) \bm{V},
\end{equation}
where $\bm{Q_d}$ is the query vector computed by $\bm{F_i^a}$, $\bm{K} \in \mathbb{R}^{C \times 1}$, $\bm{V} \in \mathbb{R}^{C \times 1}$ are projections of $\bm{E_a}$ with several linear layers, and $d_k$ denotes the dimension of keys.
Next, the output $\bm{E_d}$ is utilized to produce the weight of convolution filters.
Because $\bm{E_d}$ includes degradation and content information, therefore the generated filters can provide SR network the adaptive features $\bm{\tilde{F}_i}$ with degradation prior.
We also employ the channel-wise attention layer (CA) in each DQA module.
$\bm{E_d}$ is fed into the MLP to produce the channel-wise coefficients $\bm{c_f \in \mathbb{R}^{C}}$, then, convoluted feature  $\bm{\tilde{F}_i}$ is multiplied by $\bm{c_f \in \mathbb{R}^{C}}$ to rescale different channel components.

\begin{figure*}[t]
	\begin{tabular}[t]{c c c c c c c }
		 & HR & Bicubic & HAN &  DASR & MANet & Ours
		\\
		\multirow{3}{*}{\includegraphics[width=0.22\textwidth,height=0.22\textwidth,valign=t]{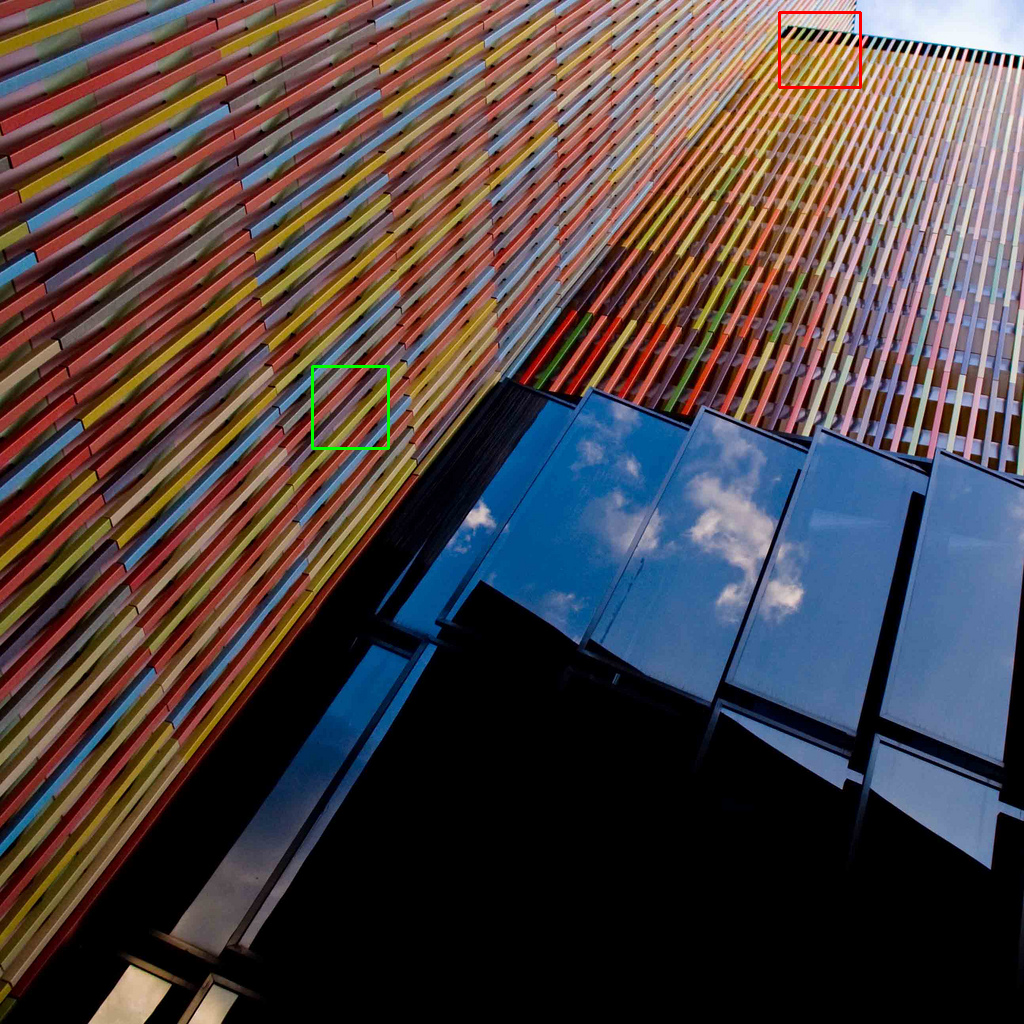}} \vspace{3pt}
		&  \includegraphics[width=.1\textwidth,height=.1\textwidth,valign=t]{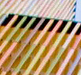} 
		&
		\includegraphics[width=.1\textwidth,height=.1\textwidth,valign=t]{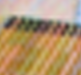} 
		& \includegraphics[width=.1\textwidth,height=.1\textwidth,valign=t]{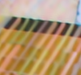} 
		& \includegraphics[width=.1\textwidth,height=.1\textwidth,valign=t]{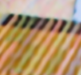} 
		& \includegraphics[width=.1\textwidth,height=.1\textwidth,valign=t]{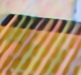}
		& \includegraphics[width=.1\textwidth,height=.1\textwidth,valign=t]{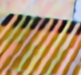}
		\\
		\vspace{-1.2mm} 
		\\
		
		& \includegraphics[width=.1\textwidth,height=.1\textwidth,valign=t]{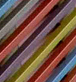} 
		&
		\includegraphics[width=.1\textwidth,height=.1\textwidth,valign=t]{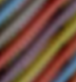} 
		&
		 \includegraphics[width=.1\textwidth,height=.1\textwidth,valign=t]{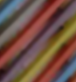} 
		& \includegraphics[width=.1\textwidth,height=.1\textwidth,valign=t]{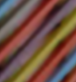} 
		& \includegraphics[width=.1\textwidth,height=.1\textwidth,valign=t]{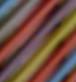}
		& \includegraphics[width=.1\textwidth,height=.1\textwidth,valign=t]{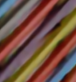}
		\\
		
		\textit{Image 23 in Urban}  & - & 18.90 dB & {\color{gray}18.42 dB} &  27.32 dB & 27.84 dB & 29.25 dB
		\\
		
		\multirow{3}{*}{\includegraphics[width=0.22\textwidth,height=0.22\textwidth,valign=t]{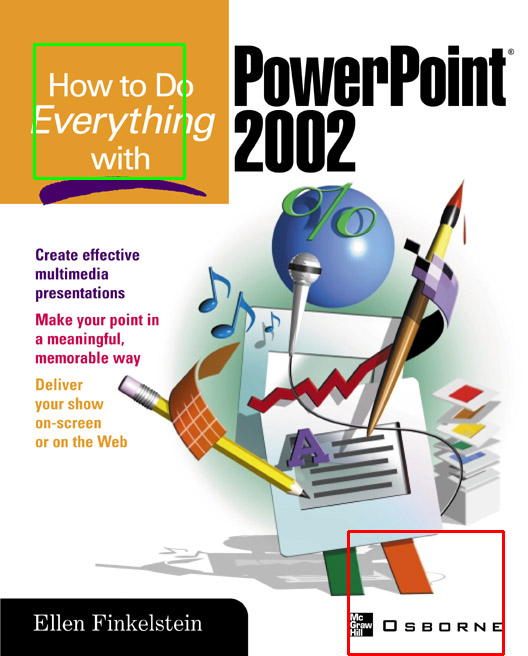}} \vspace{3pt}
		&  \includegraphics[width=.1\textwidth,height=.1\textwidth,valign=t]{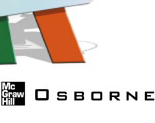} 
		&
		\includegraphics[width=.1\textwidth,height=.1\textwidth,valign=t]{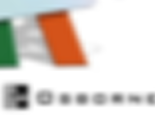} 
		& \includegraphics[width=.1\textwidth,height=.1\textwidth,valign=t]{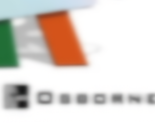} 
		& \includegraphics[width=.1\textwidth,height=.1\textwidth,valign=t]{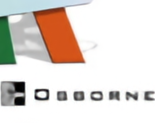} 
		& \includegraphics[width=.1\textwidth,height=.1\textwidth,valign=t]{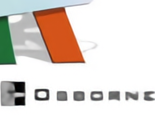}
		& \includegraphics[width=.1\textwidth,height=.1\textwidth,valign=t]{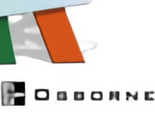}
		\\
		\vspace{-1.2mm}
		\\
		
		& \includegraphics[width=.1\textwidth,height=.1\textwidth,valign=t]{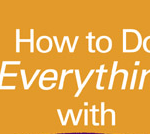} 
		&
		\includegraphics[width=.1\textwidth,height=.1\textwidth,valign=t]{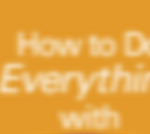} 
		&
		\includegraphics[width=.1\textwidth,height=.1\textwidth,valign=t]{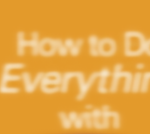} 
		& \includegraphics[width=.1\textwidth,height=.1\textwidth,valign=t]{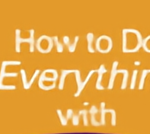} 
		& \includegraphics[width=.1\textwidth,height=.1\textwidth,valign=t]{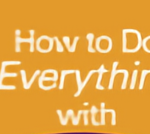}
		& \includegraphics[width=.1\textwidth,height=.1\textwidth,valign=t]{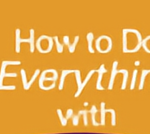}
		\\
	
		\textit{`ppt3' in Set14}  & - & 17.07 dB & {\color{gray}16.90 dB} &  26.04 dB & 26.19 dB & 26.76 dB
		
	\end{tabular}
	\vspace{-2mm}
	\caption{Visual results of different methods in Urban100 for scale factor of $4$. HAN uses a different down-sample manner causing the pixel-shifting, therefore the PSNR is low. }
	\vspace{-4mm}
	\label{fig:vis example}
\end{figure*}

\vspace{-3mm}
\subsection{Degradation Representation Learning}
To conduct the degradation representation learning in an unsupervised manner, followed by ~\cite{Wang2021Unsupervised_dasr}, we employ MoCo~\cite{he2020moco} to conduct the constrictive learning.
Specifically, given a batch of HR images $\{\bm{I_H^0}, \bm{I_H^1}, \cdots, \bm{I_H^b}\}$, we first randomly crop two patches from each image and blur down-scale them to get patch list $\bm{\hat{P}} = \{\bm{P_0}, \bm{P_1}, \cdots, \bm{P_b}\}$. 
Each $\bm{P_i}$ includes two patches $\{\bm{p_i^0}, \bm{p_i^1}\}$ which are processed by the same degradation factor.
Note that the degradation factors between $\bm{P_i}$ are different.
Conclusively, $\{\bm{p_i^0}, \bm{p_i^1}\}$ share the same content feature due to they are cropped from the same image, while $\{\bm{p_i^m}, \bm{p_j^{n}}; j\neq i, m,n \in [0,1] \}$ with $B$ different degradations are utilized to contrastively learn the degradation embedding.
As for building the large and consistent dictionaries for unsupervised learning, followed by MoCo~\cite{he2020moco}, the encoded representation of the current mini-batch is enqueued as the negative samples and the oldest are dequeued.
Moreover, this momentum-based procedure can progressively optimize the encoder reducing the influence of the domain gap between degradation space and content space.
Finally, the InfoNCE loss is used to conduct the contrastive learning:
\begin{equation}
\mathcal{L}_{cl} = \sum_i^B -\log \frac{\exp (E(\bm{p_i^1}) \cdot E(\bm{p_i^2}) / \tau)}{\sum_{j=1}^{N_{que}} \exp(E(\bm{p_i^1}) \cdot E(\bm{p_{que}^j}) / \tau)},
\label{clLoss}
\end{equation}
where $E(\cdot)$ denotes the encoder, ${N_{que}}$ denotes the length of queue, $\bm{p_{que}^j}$ is the $j^{th}$ negative sample, $\tau$ is a temperature hyper-parameter, and $B$ is the batch size.
After computing $\mathcal{L}_{cl}$, the whole loss function is defined as $\mathcal{L} = \mathcal{L}_{cl} + \mathcal{L}_{1}$, where $\mathcal{L}_{1}$ denotes the $L_1$ distance between SR result and the HR ground-truth.

\begin{figure}[t]
	\centering
	\includegraphics[width=0.8\linewidth]{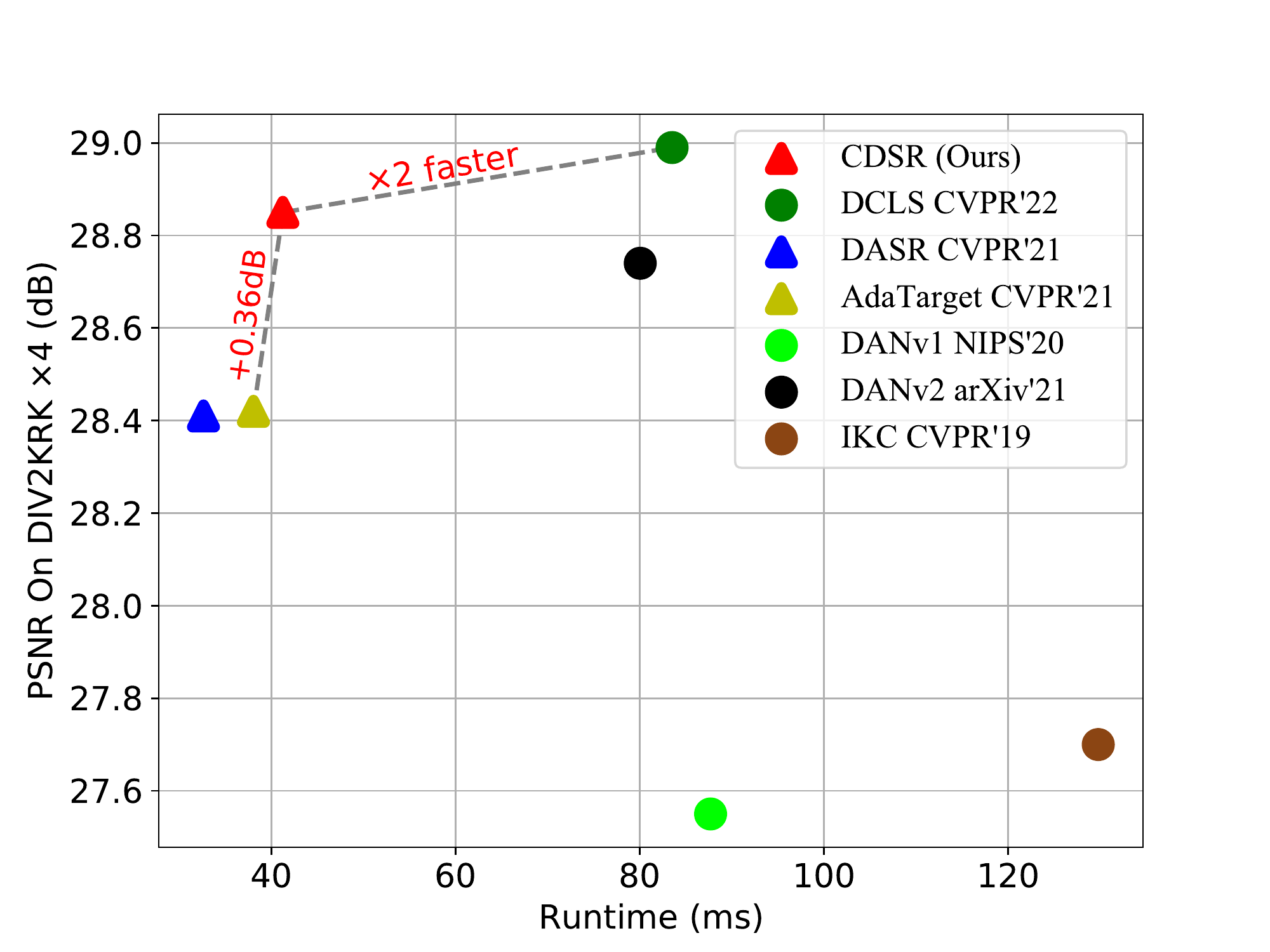}
	\vspace{-4mm}
	\caption{
		Quantitative comparison on DIV2KRK. More details are provided in the supplementary material. The triangles represent UDP methods, while the circles for SKP methods. }
	\label{fig:div2krk}
	\vspace{-4mm}
\end{figure}

\begin{figure*}[t]
	\begin{minipage}[t]{0.48\textwidth}
		\centering
		\captionof{table}{Ablation study in the proposed main components on the Urban100 dataset for $\times2$ SR.
		LPE{\scriptsize {L}} and LPE{\scriptsize {P}} denote the pixel-wise and patch-wise subnet, respectively.}
		\label{tab:ablation}
		\vspace{-3mm}
		\begin{tabular}{c|cccc|c}
			\hline
			Model        & LPE{\scriptsize {L}}  & LPE{\scriptsize {P}}  & DQA & CSC   & Urban100              \\
			\hline
			\textbf{Model 1} & \textbf{\Checkmark} & \textbf{\Checkmark} & \textbf{\Checkmark}     & \textbf{\Checkmark} & \textbf{29.57/0.8851} \\
			Model 2          & \Checkmark          & \Checkmark          & \Checkmark              & \XSolid          & 29.36/0.8828          \\
			Model 3          & \Checkmark          & \Checkmark          & \XSolid              & \Checkmark          & 29.32/0.8889          \\
			Model 4          & \Checkmark          & \XSolid          & \Checkmark              & \Checkmark          & 29.16/0.8775          \\
			Model 5          & \XSolid          & \Checkmark          & \Checkmark              & \Checkmark          & 29.44/0.8827         \\
			\hline
		\end{tabular}
		
	\end{minipage}
	\hfill
	\begin{minipage}[t]{0.48\textwidth}
		\centering
		\captionof{table}{Further analysis. SR{\scriptsize {dasr}}, SR{\scriptsize {Ours}} denotes the non-blind SR network used in DASR and our methods, respectively. E{\scriptsize {dasr}}, E{\scriptsize {manet}} denotes the encoder used in DASR, MANet.   }
		\label{tab:Further_Analysts}
		\small
		\vspace{-3mm}
		\begin{tabular}{c|c|c|c}
			\hline
			Model                                                                    & Set14        & BSD100       & Urban100     \\
			\hline
			SR{\scriptsize {dasr}} + E{\scriptsize {dasr}}                                                                 & 31.30/0.8683 & 30.46/0.8507 & 28.66/0.8654 \\
			
			SR{\scriptsize {dasr}} + LPE{\scriptsize {P}}                                                  & 31.65/0.8796 & 30.73/0.8603 & 29.10/0.8796 \\
			
			SR{\scriptsize {Ours}} + E{\scriptsize {dasr}}                                                    & 31.77/0.8777 & 30.81/0.8590 & 28.96/0.8738 \\
			
			SR{\scriptsize {Ours}} + E{\scriptsize {manet}}                                                     & 32.08/0.8825 & 30.94/0.8624 & 29.44/0.8829 \\
			SR{\scriptsize {Ours}}+LPE{\scriptsize {L}}  & 31.86/0.8787 & 30.80/0.8594 & 29.16/0.8775 \\
			\hline
			\textbf{SR{\scriptsize {Ours}}+LPE}                                                                  & \textbf{32.14/0.8841} & \textbf{31.02/0.8643} & \textbf{29.57/0.8851} \\
			\hline
		\end{tabular}
	\end{minipage}
\end{figure*}

\begin{table}[t]
	\captionof{table}{Comparison of different fusion methods for $\times2$ SR.}
	\label{tab:fuse}
	\vspace{-3mm}
	\small
	\begin{tabular}{c|c|c|c|c}
		\hline
		Method $\times2$        & Set5         & Set14        & BSD100       & Urban100     \\
		\hline
		AdaIN            & 35.76/0.9396 & 31.67/0.8778 & 30.79/0.8600 & 29.20/0.8800 \\
		DynConv          & 36.04/0.9427 & 31.71/0.8842 & 30.85/0.8678 & 29.32/0.8889 \\
		DQA       & 36.17/0.9428 & 32.14/0.8841 & 31.02/0.8643 & 29.57/0.8851 \\
		\hline
	\end{tabular}
\end{table}

\section{Experiment}
\subsection{Experimental Setup}
\noindent\textbf{Implementation Details.}
Following existing methods~\cite{Wang2021Unsupervised_dasr,manet2021,gu2019blind_ikc}, $800$ images in DIV2K~\cite{div2k} and $2,650$ images in Flickr2K~\cite{flikr2k} are collected for training. 
The training degradation default uses $21\times 21$ anisotropic Gaussian kernels and the noise level is $0$ except for the noisy $\times 4$ experiment which is set to $\mathcal{U}(0, 15)$. 
During the training stage, kernel width $\sigma_1,\sigma_2 \sim \mathcal{U}(0.175s, 2.5s)$ for scale factor $s$, and the rotation angle $\theta \sim \mathcal{U}(0, \pi)$.
The size of LR patch $\bm{\hat{P}}$ is set to $48 \times 48$ for all experiments ($\times 2, \times 3, \times 4$), therefore, the size of HR patch cropped from HR image is $96, 144, 192$, respectively.  
The batch size $B$ is $32$, \text{i.e.}, $32$ Gaussian kernels from the above ranges are randomly selected to generate LR images.
The CDSR is trained end-to-end.
The SR network employs $10$ RRDB block and DQA layers. The length of codebook $L$ is set to $1,024$. The channel number of embedding $C$ is $256$.
As for MoCo, the $\tau$ and $N_{que}$ in Eq.~\ref{clLoss} is set to $0.07$ and $8,192$, respectively.
The Adam optimizer with the momentum of $\beta_1 = 0.9$ , $\beta_2 = 0.999$  is adopted to train our network with the learning rate being initially set to $1e-4$.
the learning rate will decay by half after every $125$ epochs by the multi-step decreasing strategy.
The training process takes $500$ epoch.

\noindent\textbf{Performance Evaluation.}
Five benchmark datasets are used for evaluation: Set5~\cite{set5}, Set14~\cite{set14}, BSD100~\cite{B100}, Urban100~\cite{Urban100}, and DIV2KRK~\cite{bell2019blind_kernelGAN}. 
Followed by previous work~\cite{manet2021, luo2022DCLS}, Set5, Set14, B100 and Urban100 are degraded by $9$ different kernels: for $\times 4$ kernels are sampled from $\sigma_1,\sigma_2 \in \{1,5,9\}$ and $\theta \in \{0, \frac{pi}{4}\}$; for $\times 2$ and $\times 3$, the kernels set varies to $\sigma_1,\sigma_2 \in \{1,3,5\}$ and  $\sigma_1,\sigma_2 \in \{1,4,7\}$, respectively.
The proposed model is evaluated by PSNR and SSIM on the Y channel of the SR in YCbCr space.

\subsection{Comparison with State-of-the-Art Methods}
We conduct experiments on degradations with anisotropic Gaussian kernels and noise.
We compare CDSR with baseline models and existing blind SR models.
The results are shown in Table~\ref{tab:main_comp} and Fig.~\ref{fig:vis example}.
For fair comparisons, we retrained some of the methods under the same experimental setting.
As shown in Table~\ref{tab:main_comp}, the proposed CDSR outperforms the existing UDP models (\egs, $\times 2$ 0.8 dB better than DASR) and achieves competitive performance on PSNR and SSIM when compared with the SKP methods (MANet).
Although IKC and SRSVD struggle to estimate the accurate kernel prior by introducing the iteration kernel refinement and the adversarial network, respectively, they achieve poor performance when compared to MANet.
In particular, MANet achieves the dominant results among the SKP-based methods. 
By restricting the receptive field within a moderate range, MANet can capture the locality of degradation features without distributing.

However, these approaches occupy a large computational cost due to the iteration (IKC) and the full-size kernel maps (MANet).
Moreover, the SKP methods only consider the blur kernel, therefore, these methods suffer from severe performance drop when degradation is composed of multiple degradations (\ies, noise).
On the contrary, UDP methods are not limited to learning the kernel and attempting to represent the degradation.
KernelGAN gets the worst performance compared with other UDP methods.
The implicit kernel prediction fails to capture accurate information when applied in severe degradation.
HAN employs the attention mechanism to extract the refinement feature in the SR network, the degradation prior is ignored in this method resulting in poor performance.
DIP leverages the network prior to generate the SR images and largely improves the results.
However, DIP suffers from multiple degradations and fails to produce a suitable prior. 
DASR employs a discriminative degradation encoder by unsupervised contrastive learning and largely improves the performance.
However, the domain gap between degradation space and textual space limits the results thus performing inferior to our method.
It is worth noting that our method achieves superior performance compared with other models on multiple degradations (\ies, $\times 4$ with noise level $15$).

\begin{figure}[t]
	\begin{tabular}[t]{c@{}c}
		\hspace{-8mm}
		\includegraphics[width=.6\linewidth]{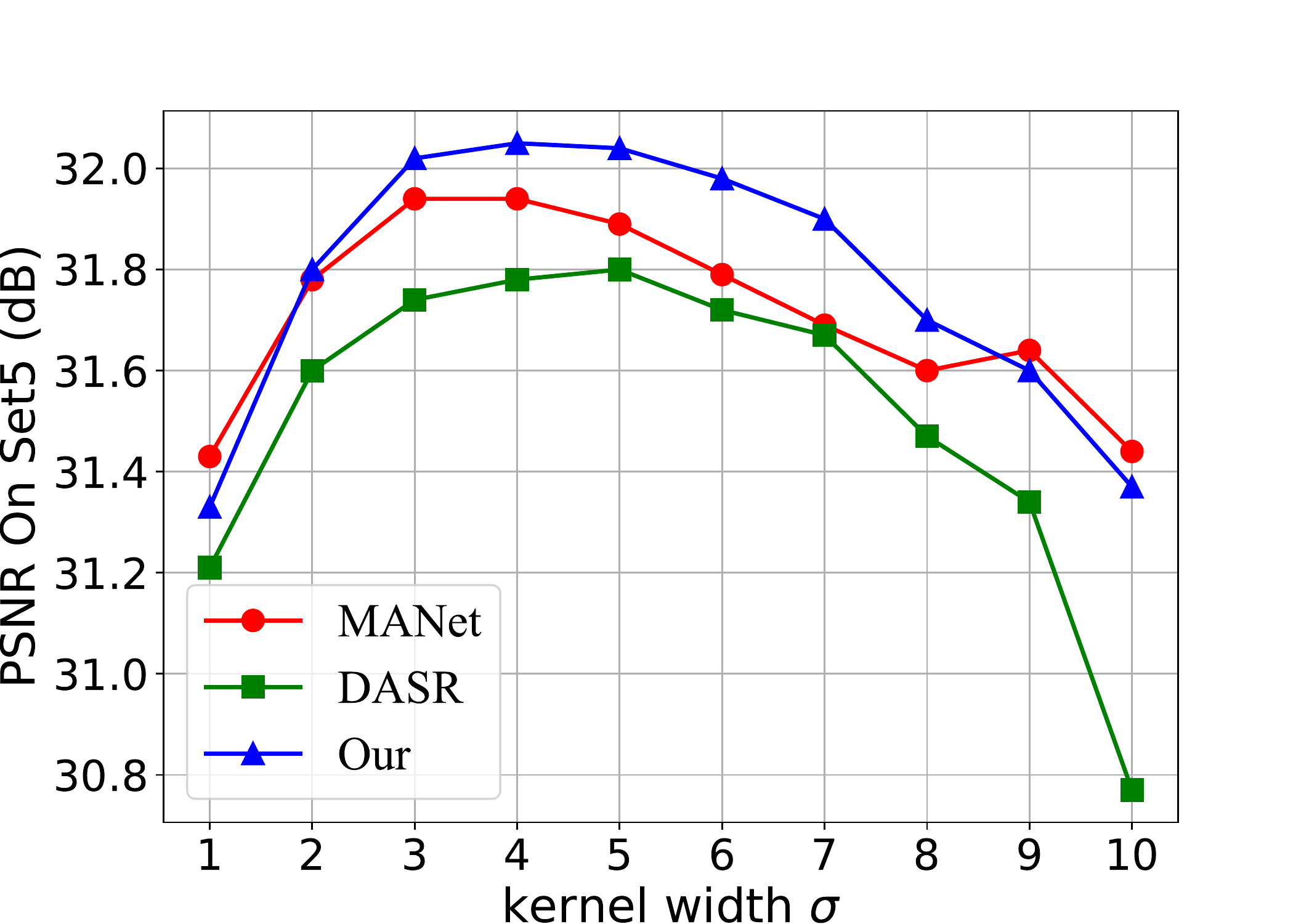}&  
		\hspace{-5mm}
		\includegraphics[width=.6\linewidth]{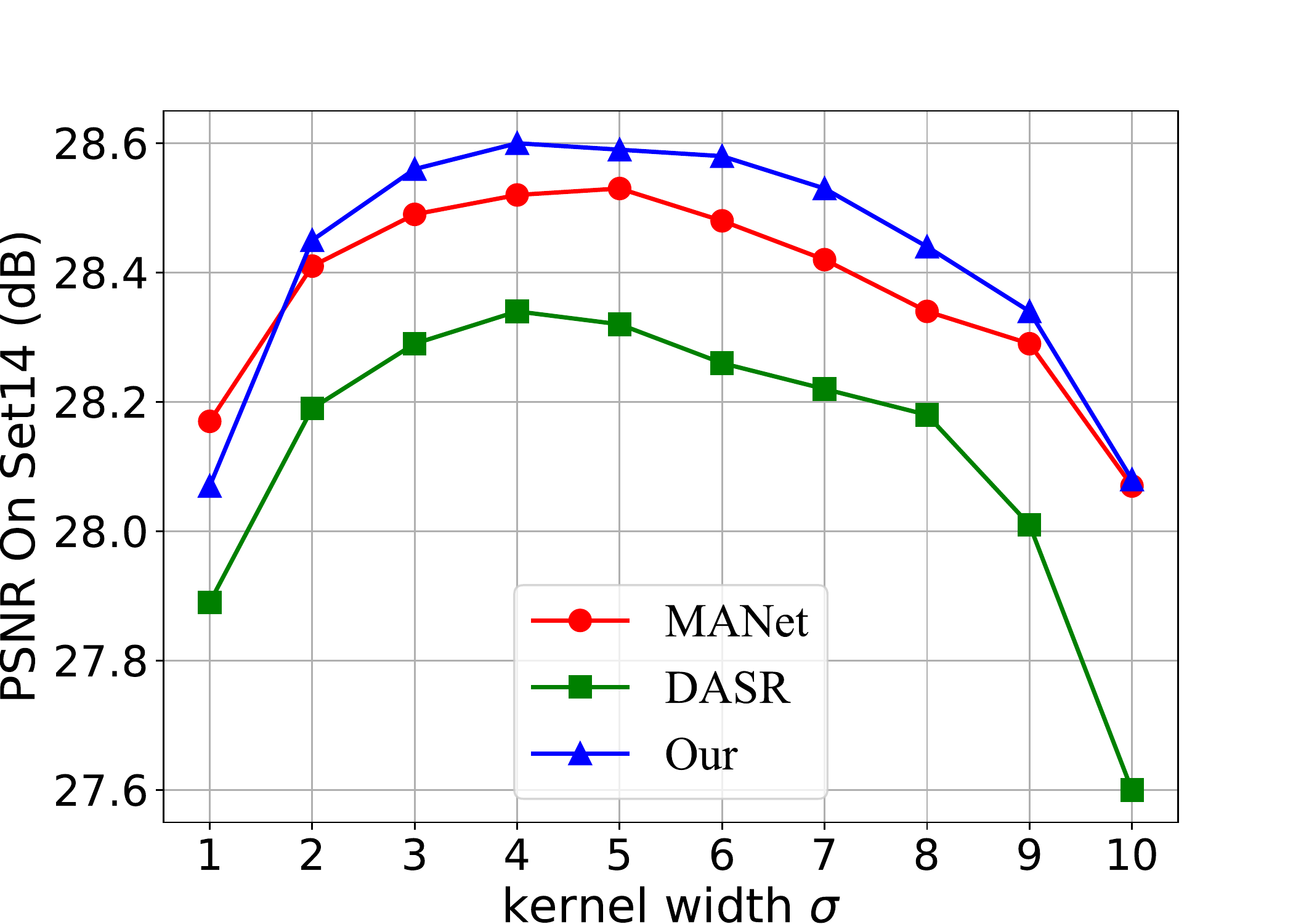}	
		\\	
	\end{tabular}
	\vspace{-4mm}
	\caption{The PSNR curves on Set5 (left) and Set14 (right) of scale factor 4. The kernel width $\sigma_1=\sigma_2=[1:10:1]$.}
	\label{fig:curve}
\end{figure}

In addition, we retrain our model with the same experimental setting as previous methods on DIV2KRK~\cite{bell2019blind_kernelGAN}.
The quantitative comparisons are shown in Fig.~\ref{fig:div2krk}.
It can be found that our proposed method can significantly outperform existing UDP methods and even shows the competitive results compared with the SKP methods.
As an improved version of IKC, DAN has achieved remarkable performance using the iteration kernel refinement.
In addition, DCLS largely improves the results by accurate estimation of kernel information.
CDSR achieves the second-best of all the methods including SKP and UDP methods by taking \textbf{13.23 M} while DCLS takes \textbf{19.05 M} parameter's amount.
For SKP methods, CDSR is almost $2\times$ faster than DCLS (Test on a single V100) and achieves the competitive results compared with it.
For UDP methods, CDSR outperforms the AdaTarget and the DASR by 0.36 dB on PSNR with similar computational cost.
To show the variable trend with the change of $\sigma$ we also provide the various curves on Set5 and Set14, as shown in Fig.~\ref{fig:curve}.

\begin{figure}[t]
	
	\begin{tabular}[t]{c@{} c@{} c@{} c}
		  & Bicubic & Model 2 & Model 3 
		\\
		\hspace{-2mm}
		\multirow{3}{*}{\includegraphics[width=0.47\linewidth,height=0.45\linewidth,valign=t]{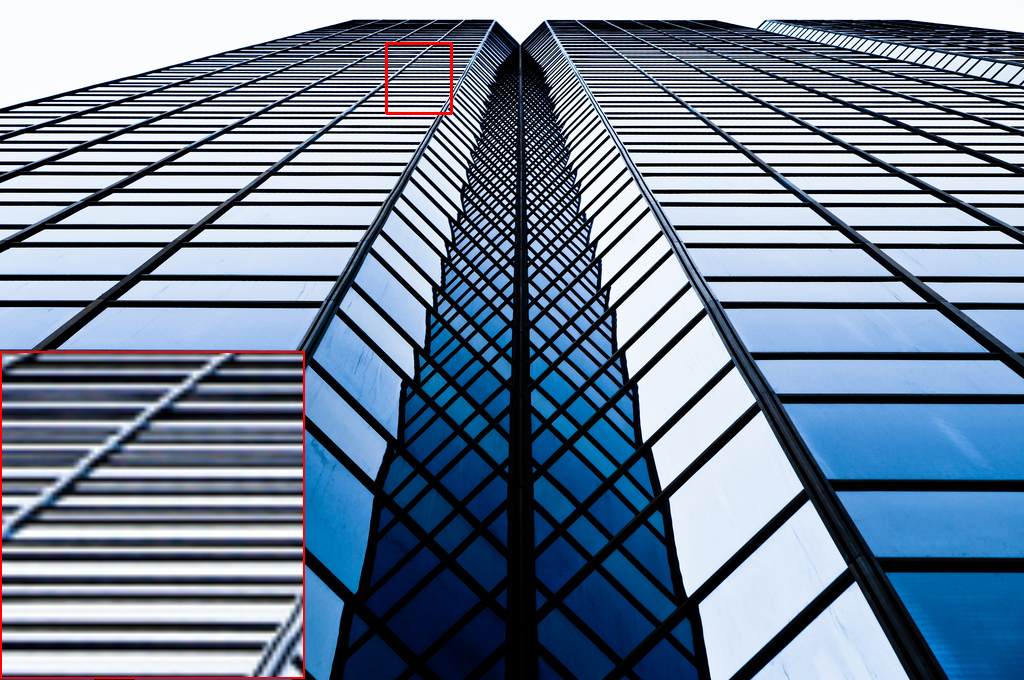}} \vspace{3pt}
		&  \includegraphics[width=.17\linewidth,height=.22\linewidth,valign=t]{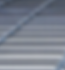} 
		& \hspace{1pt}
		\includegraphics[width=.17\linewidth,height=.22\linewidth,valign=t]{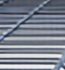} 
		& \hspace{1pt}
		\includegraphics[width=.17\linewidth,height=.22\linewidth,valign=t]{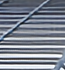} 
		\\
			
		&  \includegraphics[width=.17\linewidth,height=.22\linewidth,valign=t]{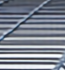} 
		& \hspace{1pt}
		\includegraphics[width=.17\linewidth,height=.22\linewidth,valign=t]{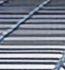} 
		& \hspace{1pt}
		\includegraphics[width=.17\linewidth,height=.22\linewidth,valign=t]{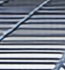} 
		\\
		\textit{urban 67}  & Model 4 & Model 5 & Model 1 \vspace{-40mm}  
		\\
		& \color{white}\hspace{-3mm}15.81 dB  & \hspace{-2mm}21.33 dB & \hspace{-2mm}22.97 dB 	\vspace{15mm}  
		\\
		& \hspace{-3mm}22.24 dB & \hspace{-2mm}22.57 dB & \hspace{-2mm}23.21 dB 	\vspace{18mm}  
		\\
		
	\end{tabular}

	\vspace{-5mm}
	\caption{Different Visualized SR results ($\times2$ ) for ablation study models on Urban100 Img 67.} 
	\label{fig:abl}
	\vspace{-5mm}
\end{figure}


\subsection{Ablation Study}
We conduct ablation studies to validate the effect of each component in our proposed method.
The quantity results are shown in Table~\ref{tab:ablation}, and the quality visualizations are shown in Fig.~\ref{fig:abl}.
All the experiments are conducted on the scale factor of $\times 2$ setting.

\noindent\textbf{Effect of Patch-wise Subnet.}
The patch-wise subnet in LPE is designed to extract the content information.
Based on the content information, we can close the domain gap between the degradation embedding space and the texture space used in the SR network.
To demonstrate its effect, we discard the patch-wise subnet and only apply the pixel-wise sub-net to extract the embedding.
In addition, other components are maintained.
Shown in Table~\ref{tab:ablation} "Model 4", the performance significantly drops from $29.57$ to $29.16$.

\noindent\textbf{Effect of Pixel-wise Subnet.}
We also test LPE without the pixel-wise subnet, denoted as Model 5.
From the quantity results, we can find that pixel-wise subnet has a moderate effect.
In other words, models with patch-wise subnet can preserve mainly SR capability.
A reasonable explanation is the $\bm{E_p}$ is able to produce a promising embedding and $\bm{E_l}$ serves as the supplementary feature. 
This is also coherent with results in Table~\ref{tab:Accuracy the Better?}: the model which only focuses on local degradation features is inferior to that on content features.

\noindent\textbf{Effect of DQA Module.}
In order to adaptively fuse the embedding prior to the SR network, the DQA module is proposed to reduce the domain gap.
To study the effect of DQA, we conduct the experiment without it, which is denoted as Model 3.
The self-attention mechanism applied in DQA can leverage the content cue to query the suitable embedding features, therefore there are fewer artifacts in the produced SR results.
As shown in Fig.~\ref{fig:abl}, the output of Model 4 suffers from the ghost artifacts, which are caused by the inconsistency of degradation embedding and SR feature.

Moreover, to demonstrate the effect of DQA, we compare this module with other fusion modules, as shown in Table~\ref{tab:fuse}.
We first test the AdaIN approach.
Specifically, instead of using Eq.~\ref{eq:sa}, the embedding $\bm{E_a}$ is fed to MLP layers to produce the scale $\beta_a \in \mathbb{R}^{C\times1}$ and the bias $\gamma_a \in \mathbb{R}^{C\times1}$.
The $\beta_a$ and $\gamma_a$ are used to refine the input features $\bm{F}_i$ by $\bm{\tilde{F}_i} = \beta_a * \bm{F}_i + \gamma_a$, where the  $\beta_a$ and $\gamma_a$ are expanded along the spatial dimension.
In addition, we also compared DQA with the standard dynamic convolution layer, shown in Table~\ref{tab:fuse}.

\noindent\textbf{Effect of Codebook.}
Intuitively, an unconstrained embedding space prefers to preserve the redundant information.
To enhance the robustness of LPE, the codebook is used to constrict the basis of embedding space.
We conduct the experiment without the codebook in LPE denoted as "Model 2".
As demonstrated in Table~\ref{tab:ablation} and Fig.~\ref{fig:abl}, the codebook can boost the performance of SR results.

\vspace{-4mm}
\subsection{Further Analysis}
Our proposed method employs a prevalent framework of blind SR, which can be summered as two steps:
(1) Degradation feature extraction; (2) embedding prior fusion in the non-blind SR network.
Therefore, the encoder and non-blind SR network are more likely to have a win-win cooperation.
CDSR takes the content information as the cue to connect them and achieve superior results. 
The experiments are conducted on the scale factor of $\times 2$ setting.

\noindent\textbf{Is Content-aware Important?}
DASR employs a naive encoder and estimates the degradation-aware embedding by contrastive learning, and the content information is learned implicitly, as shown in Table~\ref{tab:Further_Analysts}.
We replace the encoder of DASR E{\scriptsize {dasr}} by the patch-wise subnet LPE{\scriptsize{P}} to show the effect of content information (average increase of $0.33$ dB).
Besides, we replace LPE with E{\scriptsize {dasr}} to validate the effect of LPE.
Compared with DASR, the SR network with DQA can improve the output quality but is still inferior to our result.
The reason may be E{\scriptsize {dasr}} introduces the perturbation to the SR network, and destroy the cooperative relationship between LPE and DQA.

\noindent\textbf{Is Local-aware Enough?}
For the elaborate design of channel-wise relation extraction in MANet, it achieves more promising results than E{\scriptsize {dasr}} because of the moderate receptive field.
In order to study whether the performance is improved by the channel-wise mutual computation or the local receptive field, we conduct the experiments with the LPE{\scriptsize{L}} (\ies, MANet vs LPE{\scriptsize{L}}).
LPE{\scriptsize{L}} only applies several convolution layers while MANet suffers from huge computational costs.
The result shows that simply fixing the small receptive field does not share the competitive performance of MANet, but is better than E{\scriptsize {dasr}}, especially in the tough case (\egs, $29.16$ dB vs $28.96$ dB in Urban100).
It is worth noting that the performance can be significantly improved when combining the LPE{\scriptsize{L}} and LPE{\scriptsize{P}}. 


\noindent\textbf{Computation Cost and Parameters.}
For the computational cost, we study the capability of our model with different numbers of blocks in the SR network.
The inference Giga Floating-point Operations Per Second (GFlops) and the total amount of parameters are used to evaluate the computational cost.
As shown in Table~\ref{tab:compute cost}, DASR takes fewer GFlops and parameters to achieve moderate performance.
To prove the superiority is caused by the effective design rather than the computation increasing, we shrink our model by modifying the numbers of blocks used in our SR network.
When the computational cost is restricted to the same level, our proposed method is still able to outperform DASR.
In addition, the model utilizes similar computational cost to MANet and can achieve competitive performance when compared with the SOTA SKP methods. 

\begin{figure}[t]
	\begin{tabular}[t]{c@{} c @{}c}
		 & DASR & HAN 
		\\
		\multirow{3}{*}{\includegraphics[width=0.48\linewidth,height=0.41\linewidth,valign=t]{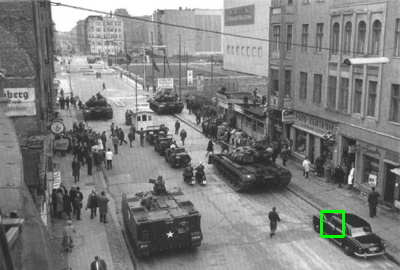}} \vspace{3pt}
		&  \includegraphics[width=.22\linewidth,height=.2\linewidth,valign=t]{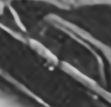} 
		& \hspace{1pt}
		\includegraphics[width=.22\linewidth,height=.2\linewidth,valign=t]{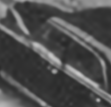} 
		
		\\		
		
		&  \includegraphics[width=.22\linewidth,height=.2\linewidth,valign=t]{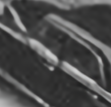} 
		& \hspace{1pt}
		\includegraphics[width=.22\linewidth,height=.2\linewidth,valign=t]{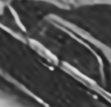} 
		
		\\
		\textit{real image}  & MANet & Ours 
		
	\end{tabular}
	\vspace{-5mm}
	\caption{Visual results of different methods on real image for $\times4$ SR. Ours provides more clear results.} 
	\label{fig:vis real example}
	\vspace{-4mm}
\end{figure}

\noindent\textbf{Performance on Real Degradation.}
To further demonstrate the effectiveness of our method, following ~\cite{Wang2021Unsupervised_dasr}, we test our model on the real images.
Visualization results are shown in Fig.~\ref{fig:vis real example}, compared with the SOTA methods, CDSR can provide more clear results, especially for the edge.
\begin{table}
	\captionof{table}{Performance and complexity for $48$$\times$$48$ image. }
	\label{tab:compute cost}
	\centering
	\vspace{-3mm}
	\begin{tabular}{c|cc|c}
		\hline
		Model   & Gflops (G) & Param. (M) & BSD100       \\
		\hline		
		DASR    & 5.97          & 5.82       & 30.46/0.8507 \\
		{\color{gray}MANet}   & {\color{gray}21.39}         & {\color{gray}9.9}        & {\color{gray}30.97/0.8651} \\
		\cdashline{1-4}[3pt/2.5pt]
		Ours (Blks=2)  & 4.93          & 5.24       & 30.72/0.8600 \\
		Ours (Blks=5)  & 10.47         & 8.24       & 30.86/0.8605 \\
		Ours (Blks=10) & 19.72         & 13.23      & 31.02/0.8643 \\
		\hline
	\end{tabular}
\end{table}

\section{Conclusion}
In this paper, we study the problem of what kind of embedding is needed for blind SR.
By deriving the truth that content information can serve as the cue for SR feature, we propose CDSR that jointly learns the content and degradation aware embedding feature for blind SR.
Specifically, the LPE is exploited to produce the degradation meanwhile preserving the content information. 
Then, the proposed DQA module can leverage the content information to adaptively query the degradation information.
Furthermore, we introduce the CSC to limit the basis of feature space and achieve end-to-end training.
Extensive experiments demonstrate that the proposed CDSR is able to achieve competitive results.


\bibliographystyle{ACM-Reference-Format}
\bibliography{sample-base}

\newpage
\appendix
Here we provide more details for the proposed CDSR.
We first give the architecture of the LPE and the DQA based non-blind SR model.
Then we provide more experiments results for CDSR.
Last, we report more quantity and quality results of CDSR.
\section{Training Details of CDSR}
The CDSR is trained end-to-end, different from the DASR~\cite{Wang2021Unsupervised_dasr} which requires to pretrain the encoder for 100 epoch, CDSR jointly train the encoder and non-blind SR in order to product more adaptive feature.
Besides, different from the MANet~\cite{manet2021} which employs complex multi-training stages, CDSR is easy to be trained.
The Adam optimizer with the momentum of $\beta_1 = 0.9$ , $\beta_2 = 0.999$  is adopted to train our network with the learning rate being initially set to $1e-4$.
the learning rate will decay by half after every $125$ epochs by the multi-step decreasing strategy.
The training process takes $500$ epoch.
The training time is about 2 days on a Tesla V100 GPU.
\subsection{Framework}
The framework details of CDSR is shown in Fig.~\ref{fig:framework}. 
CDSR is composed of two parts: (1) degradation prior encoder; (2) DQA based non-blind SR network.
\begin{figure}[!htb]
	\centering
	\includegraphics[width=\linewidth]{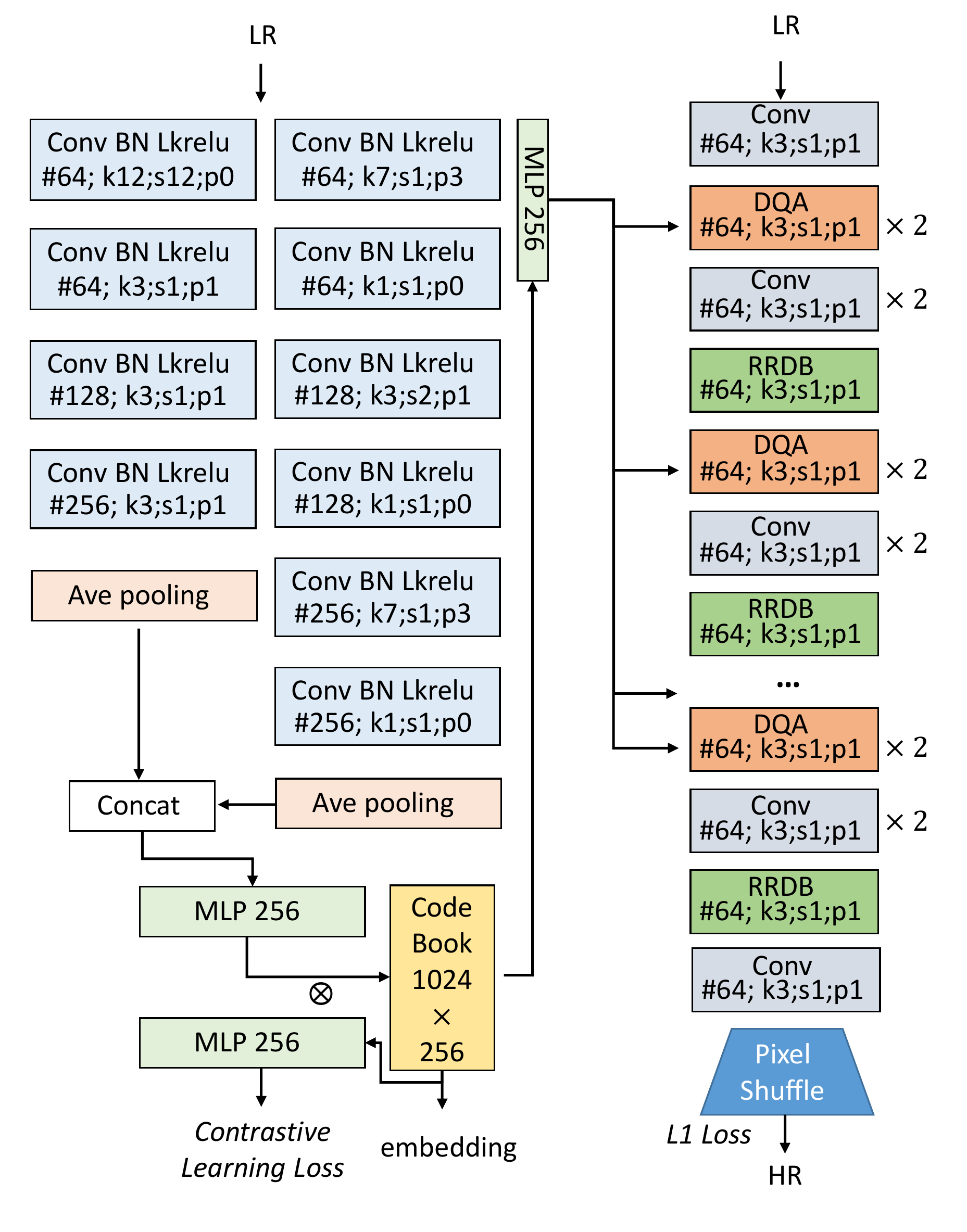}
	\vspace{-7mm}
	\caption{
		The framework of proposed CDSR. The basic block is composed by convolutional layer, batch normalization layer, leaky relu. `\#' denotes the number of filters, `k' denotes the kernel size and `p' denotes the padding size. The $otimes$ denotes the matrix multiplication.
	}
	\label{fig:framework}
\end{figure}
\begin{table}[!ht]
	\caption{Quantitative comparison on DIV2KRK. The best result is marked in {\color{red}red}. The `SK' denotes whether correspond method belongs to SKP or UDP. }
	\label{tab:DIV2KRK}
	\vspace{-3mm}
	\begin{tabular}{l|c|c|c}
		\hline
		\multicolumn{1}{c|}{\multirow{2}{*}{Method}} & \multicolumn{1}{c|}{\multirow{2}{*}{SK}} & \multicolumn{2}{c}{DIV2KRK}                      \\ \cline{3-4} 
		\multicolumn{1}{c|}{}                        & \multicolumn{1}{c|}{}                                            & \multicolumn{1}{c|}{$\times2$} & \multicolumn{1}{c}{$\times4$} \\ \hline
		{\color{gray}IKC}~\cite{gu2019blind_ikc}                                           & \Checkmark                                                                & -                       & {\color{gray}27.70/0.7668}            \\
		{\color{gray}DANv1}~\cite{huang2020unfolding}                                         & \Checkmark                                                                & {\color{gray}32.56/0.8997}            & {\color{gray}27.55/0.7582}            \\
		{\color{gray}DANv2}~\cite{luo2021endtoendDAN}                                         & \Checkmark                                                                & {\color{gray}32.58/0.9048}            & {\color{gray}28.74/0.7893}            \\
		{\color{gray}KOALAnet}~\cite{kim2021koalanet}                                      & \Checkmark                                                                & {\color{gray}31.89/0.8852}            & {\color{gray}27.77/0.7637}            \\
		{\color{gray}DCLS}~\cite{luo2022DCLS}                                         & \Checkmark                                                                & \textbf{32.75/}{\color{red}\textbf{0.9094}}            & {\color{red}\textbf{28.99/0.7946 }}           \\
		{\color{gray}Ours-SKP}                                        & \Checkmark                                                                & {\color{red}\textbf{32.76}}\textbf{/0.9050 }           & {\textbf{28.87/0.7923 }}
		\\
		
		\cdashline{1-4}[3pt/2.5pt]
		
		DBPN + Correction                             & \XSolid                                                                & 30.38/0.8717            & 26.79/0.7426            \\
		KernelGAN + ZSSR                              & \XSolid                                                                & 30.36/0.8669            & 26.81/0.7316            \\
		AdaTarget                                     & \XSolid                                                                & -                       & 28.42/0.7854            \\
		DASR                                          & \XSolid                                                                & 32.24/0.8960            & 28.41/0.7813            \\  

		Ours                                          & \XSolid                                                                & {\color{red}\textbf{32.68/0.9039}}         & {\color{red}\textbf{28.85/0.7901}}        
		\\
		\hline     
	\end{tabular}
\end{table}
\section{More Experiments}
\subsection{More details of DIV2KRK.}
Here we provide more quality comparison results for SOTA methods on DIV2KRK test set.
CDSR achieves the second-best of all the methods including SKP and UDP methods by taking \textbf{13.23 M} while DCLS takes \textbf{19.05 M} parameter's amount.
We also experiment CDSR in a SKP manner (Ours-SKP), specifically, we add a MLP with 5 layers taking the embedding as input and predict the kernel value.
Specifically, the mean absolute error (MAE) is used as the loss function to measure the difference between estimated kernels and ground-truth kernels.
Then the embedding are used in the DQA based non-blind SR.
The results are shown in Table~\ref{tab:DIV2KRK}.

\subsection{More details for experiments}
In supplementary we report the whole results for the models in our ablation study on Set5, Set14, BSD100, and Uraban100 test sets (Table~\ref{tab:abl}).
In addition, we also provide the whole results for the models in Sec 5.4 in Table~\ref{tab:further analys} and Table~\ref{tab:computational}.

\begin{table*}
\caption{More details of ablation study for $\times 2$ SR . }
\label{tab:abl}
\begin{tabular}{c|cccc|cccc}
			\hline
			Model        & LPE{\scriptsize {L}}  & LPE{\scriptsize {P}}  & DQA & CSC   & Set5         & Set14        & BSD100       & Urban100              \\
			\hline
			\textbf{Model 1} & \textbf{\Checkmark} & \textbf{\Checkmark} & \textbf{\Checkmark}     & \textbf{\Checkmark} & \textbf{36.17/0.9428} & \textbf{32.14/0.8841} & \textbf{31.02/0.8643} & \textbf{29.57/0.8851} \\
			Model 2          & \Checkmark          & \Checkmark          & \Checkmark              & \XSolid          & 36.01/0.9416          & 32.03/0.8811          & 30.90/0.8611          & 29.36/0.8828          \\
			Model 3          & \Checkmark          & \Checkmark          & \XSolid              & \Checkmark          & 36.04/0.9427          & 31.71/0.8842          & 30.85/0.8678          & 29.32/0.8889          \\
			Model 4          & \Checkmark          & \XSolid          & \Checkmark              & \Checkmark          & 35.84/0.9400          & 31.86/0.8787          & 30.80/0.8594          & 29.16/0.8775         \\
			Model 5          & \XSolid          & \Checkmark          & \Checkmark              & \Checkmark          & 35.98/0.9414          & 32.07/0.8819          & 30.89/0.8609          & 29.44/0.8827          \\
			\hline
		\end{tabular}
\end{table*}

\begin{table*}[!ht]
\caption{More details of further analysis for $\times 2$ SR. }
\label{tab:further analys}
        \begin{tabular}{c|c|c|c|c}
			\hline
			Model                                            &Set5                        & Set14        & BSD100       & Urban100     \\
			\hline
			SR{\scriptsize {dasr}} + E{\scriptsize {dasr}}                                                                 &35.30/0.9360 & 31.30/0.8683 & 30.46/0.8507 & 28.66/0.8654 \\
			
			SR{\scriptsize {dasr}} + LPE{\scriptsize {P}}               &35.55/0.9290                                   & 31.65/0.8796 & 30.73/0.8603 & 29.10/0.8796 \\
			
			SR{\scriptsize {Ours}} + E{\scriptsize {dasr}}              &35.90/0.9406                                      & 31.77/0.8777 & 30.81/0.8590 & 28.96/0.8738 \\
			
			SR{\scriptsize {Ours}} + E{\scriptsize {manet}}             &35.94/0.9409                                        & 32.08/0.8825 & 30.94/0.8624 & 29.44/0.8829 \\
			SR{\scriptsize {Ours}}+LPE{\scriptsize {L}}  &35.84/0.9400 & 31.86/0.8787 & 30.80/0.8594 & 29.16/0.8775 \\
			\hline
			\textbf{SR{\scriptsize {Ours}}+LPE}                         &\textbf{36.17/0.9428}                                       & \textbf{32.14/0.8841} & \textbf{31.02/0.8643} & \textbf{29.57/0.8851} \\
			\hline
		\end{tabular}
\end{table*}

\begin{table*}[!ht]
\caption{More details of computational cost for $\times 2$ SR. }
\label{tab:computational}
        \begin{tabular}{c|cc|cccc}
    		\hline
    		Model   & Gflops (G) & Param. (M) &Set5    & Set14        & BSD100       & Urban100       \\
    		\hline		
    		DASR    & 5.97          & 5.82   &35.30/0.9360 & 31.30/0.8683   & 30.46/0.8507 & 28.66/0.8654 \\
    		{\color{gray}MANet}   & {\color{gray}21.39}         & {\color{gray}9.9}   & {\color{gray}35.98/0.9420}  & {\color{gray}31.95/0.8845}   &  {\color{gray}30.97/0.8651}  & {\color{gray}29.61/0.8880}\\
    		\cdashline{1-7}[3pt/2.5pt]
    		Ours (Blks=2)  & 4.93          & 5.24       &35.60/0.9401 &31.69/0.8812 & 30.72/0.8600 &28.91/0.8737\\
    		Ours (Blks=5)  & 10.47         & 8.24       &35.92/0.9407 &31.90/0.8809 & 30.86/0.8605 &29.18/0.8784\\
    		Ours (Blks=10) & 19.72         & 13.23      &36.17/0.9428 &32.14/0.8841 &31.02/0.8643 &29.57/0.8851\\
    		\hline
	    \end{tabular}
\end{table*}

\begin{table*}[!ht]
    \caption{The effect of codebook with different length for $\times 2$ SR.}
    \label{tab:cbL}
    \begin{tabular}{c|cccc}
    \hline
    CodeBook Length & Set5                  & Set14                 & BSD100                & Urban100              \\
    \hline
    \textbf{1024}   & \textbf{36.17/0.9428} & \textbf{32.14}/0.8841 & \textbf{31.02/0.8643} & 29.57/0.8851 \\
    2048            & 36.11/0.9424          & 32.14\textbf{/0.8844 }         & 30.96/0.8631          & \textbf{29.59/0.8865}  
    \\
    \hline
    \end{tabular}
\end{table*}

\begin{table*}[!htb]
\caption{The impact of pretraining the encoder for $\times 2$ SR. }
\label{tab:pretrain}
    \begin{tabular}{c|cccc}
    \hline
    Pretrain Encoder  & Set5                  & Set14                 & BSD100                & Urban100              \\
    \hline
    100 epcoh        & 36.08/0.9417          & 32.05/0.8822          & 30.91/0.8611          & 29.44/0.8826          \\
    \textbf{0 epoch} & \textbf{36.17/0.9428} & \textbf{32.14/0.8841} & \textbf{31.02/0.8643} & \textbf{29.57/0.8851}
    \\
    \hline
    \end{tabular}

\end{table*}

\subsection{About the length of codebook.}
We also conduct the experiments of different length of codebook.
As shown in Table~\ref{tab:cbL}, the larger codebook can not produce the competitive results with that of original length.
The Codebook-based Space Compress module (CSC) is designed to limit the basis of feature space, thus reducing redundancy and mitigating the domain gap between degradation and content spaces.
Once enlarging the codebook, the redundancy can not be suppressed and the larger codebook will reduce the constraint of space basis.

\subsection{Pretrained encoder.}
Both DASR and MANet employ a complex training strategies to train their models: pretrain the encoders (DASR); pretrain the kernel predictor (MANet).
However, CDSR is trained end-to-end and we find that the pretrain encoder will drops the performance of CDSR. 
We conduct the experiment that firstly pretrain 100 epoch of the encoder and then train the encoder and SR network together, the results is shown in Table~\ref{tab:pretrain}.
An explanation is that, the joint training let the encoder to produce a more adaptive embedding and easy to learn the content and degradation aware embedding.

\end{document}